\documentclass{article}
\usepackage{graphicx} % 
\usepackage{amssymb}   
\usepackage{comment}
\usepackage{amsmath}
\usepackage{authblk}
\usepackage{xcolor}
\usepackage{hyperref}
\usepackage{subcaption}
\usepackage{booktabs}
\usepackage{multirow}
\usepackage{microtype}
\usepackage{float}
\usepackage[T1]{fontenc}
\usepackage[utf8]{inputenc}
\usepackage{orcidlink}

\usepackage[a4paper,
            left=3cm,
            right=3cm,
            top=1.5cm,
            bottom=1.5cm]{geometry}

\usepackage[
    style=numeric,
    sorting=none,
    doi=true,
    url=false
]{biblatex}
\title{Pretty Good Measurement for Radiomics: A Quantum-Inspired Multi-Class Classifier for Lung Cancer Subtyping and Prostate Cancer Risk Stratification}

\author{%
Giuseppe Sergioli$^{1}$\orcidlink{0000-0002-3650-5858},
Carlo Cuccu$^{1}$\orcidlink{0000-0002-4519-1854},
Giovanni Pasini$^{2,3}$\orcidlink{0000-0002-8750-0731},
Alessandro Stefano$^{2,3}$\orcidlink{0000-0002-7189-1731},
Giorgio Russo$^{2,3}$\orcidlink{0000-0003-1493-1087}, Andrés Camilo Granda Arango$^{1}$\orcidlink{0000-0002-5255-2507}, 
Roberto Giuntini$^{1}$\orcidlink{0000-0003-2078-0297}%
}

\date{}

\affil{$^{1}$\href{https://ror.org/003109y17}{Università degli Studi di Cagliari}, Via Is Mirrionis 1, Cagliari, 09123, Italy}
\affil{$^{2}$Institute of Bioimaging and Complex Biological Systems - \href{https://ror.org/04zaypm56}{National Research Council} (IBSBC - CNR), Cefalù, 90015, Italy}
\affil{$^{3}$National Laboratory of South, \href{https://ror.org/02pq29p90}{National Institute for Nuclear Physics}(LNS-INFN), Catania, 95125, Italy}

\begin{document}

\maketitle

\begin{abstract}
We investigate a quantum-inspired approach to supervised multi-class classification based on the \emph{Pretty Good Measurement} (PGM), viewed as an operator-valued decision rule derived from quantum state discrimination. The method associates each class with an encoded mixed state and performs classification through a single POVM construction, thus providing a genuinely multi-class strategy without reduction to pairwise or one-vs-rest schemes. In this perspective, classification is reformulated as the discrimination of a finite ensemble of class-dependent density operators, with performance governed by the geometry induced by the encoding map and by the overlap structure among classes.

To assess the practical scope of this framework, we apply the PGM-based classifier to two biomedical radiomics case studies: histopathological subtyping of non-small-cell lung carcinoma (NSCLC) and prostate cancer (PCa) risk stratification. The evaluation is conducted under protocols aligned with previously reported radiomics studies, enabling direct comparison with established classical baselines. The results show that the PGM-based classifier is consistently competitive and, in several settings, improves upon standard methods. In particular, the method performs especially well in the NSCLC binary and three-class tasks, while remaining competitive in the four-class case, where increased class overlap yields a more demanding discrimination geometry. In the PCa study, the PGM classifier remains close to the strongest ensemble baseline and exhibits clinically relevant sensitivity--specificity trade-offs across feature-selection scenarios.

These findings support the relevance of PGM-based quantum-inspired decision rules as a mathematically well-motivated and practically viable extension of quantum state discrimination techniques to genuinely multi-class learning problems and provide further evidence of their applicability in high-dimensional biomedical settings.
\end{abstract}

\section{Introduction}

In recent years, quantum computing has progressively moved beyond a purely programmatic phase and has become an area of research and development in which theoretical results, device engineering, and experimentation advance jointly \cite{Aerts_2016, AERTS2009314}. Today, hardware platforms can execute non-trivial sequences of quantum operations, grounded in formally characterized universal gate structures \cite{DallaChiara2009}, and produce experimental data that are useful both for testing foundational ideas and for assessing, in a controlled way, the performance and limitations of the available systems.

At the same time, it is widely acknowledged that the path toward fully reliable, large-scale quantum machines remains demanding. The effects of noise and the fragility of quantum states impose stringent constraints, and many of the most ambitious promises depend on further progress in stability, control, and the ability to reduce errors. Nonetheless, the overall trajectory is clear: quantum computing is taking shape as a mature field, in which long-term objectives and medium-term applications coexist, and in which a realistic assessment of what is possible today proceeds hand in hand with the definition of future prospects \cite{Aerts_2017, aerts2021modeling}.

Against this background, advances in quantum computing have made it natural to ask not only whether and when quantum devices will be able to offer an advantage over classical computation, but also in which application domains such an advantage (or even simply a practical benefit) might emerge in a more plausible and measurable way \cite{Aerts2025_ConceptualityI, aerts2021modeling, aerts2017testing}. In particular, the availability of increasingly controllable experimental platforms has encouraged a shift in perspective: from the mere demonstration of principles and theoretical prototypes to the search for contexts in which real information-processing procedures can be reconsidered in light of the quantum resources currently accessible, including alternative logical and algebraic frameworks inspired by quantum computational structures \cite{DALLACHIARA201894, Holik2018, Holik2019}, while still taking into account the limitations that remain (noise, stability, scalability).

Among these contexts, machine learning has emerged as an almost inevitable candidate. On the one hand, machine learning originates and develops precisely in response to the need to extract structure and information from large volumes of data, in settings where computational cost and model complexity rapidly become dominant. On the other hand, quantum computing has been conceived, since its original motivations, as a paradigm capable of treating more efficiently certain classes of computationally demanding problems.

The convergence of these two trajectories has led to the emergence of quantum machine learning (QML) as an autonomous area: a research program that aims to reformulate certain learning procedures in a natively quantum language, leveraging specific physical resources of quantum computation. In particular, the possibility of expressing crucial subroutines in terms of unitary operators and controlled quantum dynamics sustains the expectation that, at least in principle and for certain classes of tasks, such reformulations may translate into concrete and relevant computational speed-ups with respect to classical implementations, especially when the latter rapidly become prohibitive in terms of complexity and resources.

However, the application of quantum computing to machine learning does not exhaust the ways in which quantum theory can be put to work in this context. Alongside the properly quantum line---in which learning procedures are reformulated to be executed on quantum devices---a second strand has developed, often referred to as quantum-inspired (or quantum-like) machine learning. In this case, the inspiration comes from concepts and tools of quantum information theory, but the algorithms remain formulated in classical language and are executable on conventional computers. It follows that the objective is not primarily the speed-up linked to quantum hardware, but rather the introduction of conceptual schemes and design criteria that can guide the construction of alternative and potentially more effective learning methods.

A particularly natural source of inspiration, already discussed in the literature, is quantum state discrimination. In quantum theory, discriminating states means rigorously formalizing the distinguishability among physical preparations and deriving optimal decision procedures under conditions of unavoidable uncertainty. In the quantum-inspired setting, this perspective suggests reinterpreting the classification problem as a ``discrimination'' task among class representatives, adopting decision criteria motivated by the notion of distinguishability. Along this line, the expected benefit does not concern so much a reduction in computational complexity, but rather the possibility of achieving good levels of accuracy through decision rules that inherit, in classical form, the discriminative intuition characteristic of the quantum framework.

From this perspective, interest in quantum-inspired machine learning does not remain confined to a conceptual proposal: some early lines of research have in fact shown how a binary classifier inspired by state discrimination can be transferred in a natural way to concrete problems \cite{sergioli2017quantum, Sergioli2019}, where the quality of the decision (more than computational efficiency alone) is a central requirement. In the biomedical domain, this requirement is particularly stringent: data are often heterogeneous and noisy, and class separation may depend on weak signals or on patterns that are not immediately evident.

Against this background lie the first attempts to apply the quantum-inspired classifier to practical cases. On the one hand, the approach has been explored on clinical data in a pulmonary context, for instance for classification tasks related to survival stratification in patients with idiopathic pulmonary fibrosis, where discrimination among classes assumes a directly interpretive relevance.
On the other hand, image analysis has been considered in laboratory procedures---for example in supporting the evaluation of clonogenic assays---where the crucial task is a binary discrimination (colony vs.\ background) that affects the final quantification \cite{Sergioli2021}.

The experiments discussed so far had an intrinsically binary setting, because they were based on a quantum-inspired classifier derived directly from the Helstrom measurement, which arises as the optimal (minimum-error) solution to the problem of discriminating between two states: in machine learning terms, this translates into a natural device for two-class classification.
\textcolor{black}{More explicitly, given a binary ensemble $\{(p_1,\rho_1),(p_2,\rho_2)\}$, the Helstrom measurement is the POVM that minimizes the average probability of misclassification. It is obtained from the spectral decomposition of the Helstrom operator $\Delta := p_1\rho_1 - p_2\rho_2$: one assigns the first class on the positive eigenspace of $\Delta$ and the second class on its negative eigenspace (with a conventional treatment of any zero eigenspace). In this sense, the Helstrom construction provides a mathematically natural and decision-theoretically optimal rule for two-class discrimination. At the same time, precisely because this construction is intrinsically binary, its extension to problems with more than two labels is not immediate and typically requires decomposition strategies; this is one of the motivations for moving to the PGM framework in the genuinely multi-class setting.} In the presence of more classes, such an approach therefore tends to require decomposition strategies (e.g., one-vs-one or one-vs-rest), with a consequent multiplication of comparisons and required resources.

More recently, however, a quantum-inspired classifier natively $n$-ary has been proposed, which addresses multi-class classification directly, without reducing it to a collection of binary problems \cite{giuntini2023quantum, Giuntini_2023}. The guiding idea remains that of quantum state discrimination, but the extension to the general case is obtained by adopting the \emph{Pretty Good Measurement} (PGM), i.e., a systematic procedure to construct a measurement (POVM) \textcolor{black}{for discriminating an arbitrary finite set of states whose average success probability is provably close to optimal; more precisely, if $P_{\mathrm{opt}}$ denotes the optimal average success probability for the same ensemble, and $P_{\mathrm{PGM}}$ denotes the average success probability achieved by the PGM for that ensemble, then the PGM satisfies $P_{\mathrm{PGM}} \geq P_{\mathrm{opt}}^{2}$.}
This shift is conceptually relevant: on the one hand, it makes the treatment of datasets with multiple labels ``natural''; on the other hand, precisely because the method is formulated in the language of quantum measurement theory, it also suggests possible implementations on quantum hardware (at least in principle), in addition to immediate execution on conventional computers in a quantum-inspired form.

In an applicative perspective, this opens in a particularly direct way to the possibility of addressing multi-class biomedical problems, where classification rarely reduces to a simple ``yes/no'' and where, instead, it is often crucial to distinguish among multiple conditions, phenotypes, or levels of severity within the same clinical or experimental framework.

In this work, we consider two biomedical scenarios already well characterized in the literature and based on imaging and radiomics data. The first concerns the phenotyping of histopathological subtypes of \emph{Non-Small-Cell Lung Carcinoma} (NSCLC) from multicenter Computed Tomography (CT) images \cite{Pasini2023}, where the clinical-informative goal is to distinguish among multiple subtypes (e.g., Adenocarcinoma, Squamous Cell Carcinoma, Large Cell Carcinoma, and a ``Not Otherwise Specified'' class), in a context in which acquisition variability and pipeline choices substantially affect the difficulty of the task.
The second scenario concerns the stratification of risk in prostate carcinoma from images of \emph{Prostate-Specific Membrane Antigen} (PSMA) and positron emission Tomography (PET/CT)\cite{Pasini2025}, in which the basic setting is aimed at separating grouping of patients according to clinically significant risk classes, with the aim of supporting diagnostic and therapeutic decisions in a non-invasive way.

The objective of the article, in continuity with what has been discussed above about the transition from binary schemes to a natively $n$-ary classifier inspired by the PGM, is to show how this approach can be applied in a multi-class mode to the datasets and contexts considered in these two works, directly treating the multi-class problem (without one-vs-one or one-vs-rest strategies) and discussing in a transparent way the methodological implications and the empirical behavior on the biomedical data under examination.

The paper is organized as follows. First, to make the presentation self-contained, we provide a concise but technically explicit account of the PGM-based discrimination protocol as it is used for supervised classification. We then summarize, at a qualitative level but with the essential experimental details, the two reference applications that serve as our clinical and methodological benchmarks, namely prostate cancer and lung carcinoma. The second part of the paper is devoted to our main contribution: we report the results of applying the PGM-based quantum-inspired classifier to these two medical-imaging radiomics datasets, under evaluation protocols chosen to enable a meaningful comparison with baselines previously obtained with classical methods. The paper closes with some final remarks.

\section{The Pretty Good Measurement (PGM) classifier}
\label{sec:pgm-classifier}

In this section we summarize the multi-class quantum-inspired classifier based on the PGM, presenting the construction in a self-contained form while keeping the notation close to standard supervised learning. 

Let $L=\{1,\dots,\ell\}$ be the set of class labels, and let a training dataset be a finite set
\begin{equation}
S_{\mathrm{tr}}=\{(x_1,\lambda_1),\dots,(x_m,\lambda_m)\}, \qquad \lambda_j\in L,
\end{equation}
where \textcolor{black}{$m$ is the cardinality of the training set and} each object is represented by a feature vector $x_j\in\mathbb{C}^d$ (the real-valued case is included as a special instance). \textcolor{black}{Here and throughout the paper, we use $x_j$ to denote elements of the training set, while $x$ denotes a generic input (e.g., at inference time).}

We represent multi-class prediction by assigning a score to each class. Concretely,
we specify a scoring function
\[
f : \mathbb{C}^d \to [0,1]^\ell,\qquad 
f(x)=\bigl(f_1(x),\ldots,f_\ell(x)\bigr),
\]
and define the associated classifier by selecting the label with the highest score:
\[
Cl_f(x):=\min\Bigl\{i\in L:\ f_i(x)=\max_{1\le k\le \ell} f_k(x)\Bigr\},
\]
where, in case of ties (multiple maximizers), we choose the smallest index. As it will be shown, in the
PGM classifier the scores are produced by a measurement rule and therefore admit
a direct probabilistic interpretation.

The first quantum-inspired step consists in fixing an encoding (feature map) that associates to each input vector $x$ a density operator $\rho_x$ on a finite-dimensional Hilbert space $H\simeq\mathbb{C}^n$:
\begin{equation}
x \longmapsto \rho_x \in \mathcal{D}(H).
\end{equation}

In previous works \cite{sergioli2017quantum, Sergioli2019} different kinds of encodings have been introduced (\emph{amplitude} and \emph{stereographic} encodings).

%The training set thus induces encoded patterns $(\rho_{x_j},\lambda_j)$. For each class $i\in L$, define the class-conditioned subset $S^{i}_{\mathrm{tr}}:=\{x_j:\lambda_j=i\}$ and summarize it by a class representative in operator form. 

The encoding turns the classical training set
into its encoded counterpart
\[
\widetilde S_{\mathrm{tr}}=\{(\rho_{x_j},\lambda_j)\}_{j=1}^{m}.
\]

\textcolor{black}{For each class $i \in L$, let
$
S_{\mathrm{tr}}^{i} := \left\{\, x_j \in \mathbb{C}^d : j \in \{1,\dots,m\},\ \lambda_j = i \,\right\},
$
and define the corresponding family of encoded states as
$
\widetilde S^i_{tr}=\{\rho_{x_j}: x_j\in S^{i}_{\mathrm{tr}}\}.
$ Intuitively, $S_{\mathrm{tr}}^{i}$ collects all training inputs belonging to class $i$, while $\widetilde{S}_{\mathrm{tr}}^{i}$ is the corresponding set of their encoded quantum states.}

Now we associate to each class \(i\) a single density operator that plays
the role of a class representative.

This representative is taken to be the \emph{quantum centroid}, i.e., the uniform average of the encoded training states in that class,
\begin{equation}
\rho^{(i)}:=\frac{1}{|\widetilde S^{i}_{\mathrm{tr}}|}\sum_{x_j\in S^{i}_{\mathrm{tr}}}\rho_{x_j},\qquad i\in L.
\end{equation}
Even when the encoding produces pure states, the centroid $\rho^{(i)}$ is generally mixed, and its geometry need not coincide with that of classical centroids in feature space; this is one of the mechanisms by which the induced decision rule can depart from standard nearest-mean classifiers.

To turn these class representatives into a multi-class prediction rule, one uses a POVM-induced scoring function. A POVM with outcomes in $L$ is a family of positive semidefinite operators $\{M_i\}_{i=1}^{\ell}$ on $H$ such that $\sum_{i=1}^{\ell}M_i=I$. Given a POVM and an encoded input $\rho_x$, Born's rule produces scores
\begin{equation}
f_i(x):=\mathrm{tr}(M_i\rho_x),\qquad i\in L,
\end{equation}
which satisfy $f_i(x)\in[0,1]$ and $\sum_{i=1}^{\ell}f_i(x)=1$. 

The PGM specifies a canonical way to construct such a POVM from an ensemble of hypotheses. Consider the ensemble of class representatives with priors
\begin{equation}
\mathcal{R}=\{(p_1,\rho^{(1)}),\dots,(p_\ell,\rho^{(\ell)})\},\qquad p_i>0,\ \sum_{i=1}^{\ell}p_i=1,
\end{equation}
and define the average (mixture) state
\begin{equation}
\sigma:=\sum_{i=1}^{\ell}p_i\rho^{(i)}.
\end{equation}
The PGM elements are first defined on the support of $\sigma$ by
\begin{equation}
E_i:=\sigma^{-1/2}\,p_i\rho^{(i)}\,\sigma^{-1/2},\qquad i\in L,
\end{equation}
where $\sigma^{-1}$ denotes the Moore--Penrose pseudoinverse and $\sigma^{-1/2}$ its positive square root. In general $\sum_{i=1}^{\ell}E_i=P_{\mathrm{im}(\sigma)}$, the projection onto the image of $\sigma$, and one obtains a proper POVM by completing on the kernel:
\begin{equation}
F_i:=E_i+\frac{1}{\ell}P_{\ker(\sigma)},\qquad i\in L.
\end{equation}
Then $\sum_{i=1}^{\ell}F_i=I$, and the family $F=\{F_i\}_{i=1}^{\ell}$ defines a valid measurement. The corresponding scoring function and classifier are obtained by setting
\begin{equation}
f_i(x):=\mathrm{tr}(F_i\rho_x),\qquad \mathrm{Cl}_{\mathrm{PGM}}(x):=\min\Bigl\{\, i\in L : \mathrm{tr}(F_i\rho_x)=\max_{1\le k\le \ell}\mathrm{tr}(F_k\rho_x)\Bigr\}.
\end{equation}
In the absence of externally imposed class priors, one may adopt uniform priors $p_i=1/\ell$, so that the measurement is determined entirely by the geometry of the class centroids through $\sigma$. A key methodological point is that this construction is \emph{intrinsically multi-class}: the decision rule arises from a single $\ell$-outcome POVM and does not require one-vs-rest or one-vs-one decompositions.

An optional extension increases expressive power by lifting the representation via tensor copies. For $n\in\mathbb{N}_{+}$ one encodes an input as $\rho^{(n)}_x:=\rho_x^{\otimes n}$ and defines class representatives as
\begin{equation}
\rho^{(n)}_{(i)}:=\frac{1}{|S^{i}_{\mathrm{tr}}|}\sum_{x_j\in S^{i}_{\mathrm{tr}}}\rho_{x_j}^{\otimes n},
\end{equation}
noting that, in general, $\rho^{(n)}_{(i)}\neq(\rho^{(i)})^{\otimes n}$. The PGM construction is then repeated on the ensemble $\{(p_i,\rho^{(n)}_{(i)})\}_{i=1}^{\ell}$, yielding POVM elements $\{F^{(n)}_i\}_{i=1}^{\ell}$ and scores $f^{(n)}_i(x)=\mathrm{tr}(F^{(n)}_i\rho_x^{\otimes n})$. Empirically, increasing $n$ may improve performance at the cost of higher computational load, since training requires forming $\sigma$ and computing $\sigma^{-1/2}$ in the enlarged operator space \cite{giuntini2023quantum, Giuntini_2023}. Once the measurement is fixed, inference reduces to evaluating the traces $\mathrm{tr}(F_i\rho_x)$ (or their $n$-copy counterparts) and applying the argmax decision rule.

To further improve the classifier's accuracy, in~\cite{sergioli2017quantum} we showed that two additional choices can be tuned already at the preprocessing stage. The first concerns the \emph{encoding}, namely the map that associates each real feature vector \(x\) with a quantum state \(\rho_x\). There are many admissible ways to construct such an encoding \(x \mapsto \rho_x\) (e.g., via a direct real-vector representation or via a density-operator representation), and the most effective choice can be dataset-dependent. The second concerns the introduction of a multiplicative \emph{rescaling factor} \(\alpha>0\), applied to all feature vectors before encoding (i.e., \(x \mapsto \alpha x\)). Empirically, this global rescaling can yield an additional gain in accuracy, which is consistent with the fact that the classification procedure takes place in the geometric state space of quantum states, where distances and overlaps depend on the chosen scale of the embedded data. Accordingly, an application of the PGM classifier can include a grid-search over the main hyperparameters discussed above, namely the encoding scheme, the number of copies, and the rescaling factor.

\textcolor{black}{
From a computational viewpoint, the PGM classifier is not trained through an iterative optimization procedure. Once the encoding is fixed, training consists of constructing the class centroids in the encoded space, forming the mixture operator $\sigma$, and computing the corresponding PGM operators. Therefore, the notion of training epochs does not apply here. The dominant cost of the training phase is the computation of the (pseudoinverse) square root of $\sigma$, which is the only step that scales superlinearly with the dimension of the encoded space; this dimension grows with the number of tensor copies, which makes the copy count the most influential parameter for runtime and memory. Inference, in contrast, reduces to a finite set of trace evaluations followed by an $\arg\max$ decision rule, and is therefore inexpensive once the measurement is fixed.}

\section{Clinical Imaging Baselines: Radiomics Pipelines and Reported Performance on Lung Carcinoma and Prostate Cancer Datasets}

{\textcolor{black}{Radiomics provides a methodological bridge between medical imaging and data‑driven learning by transforming images into high‑dimensional quantitative representations \cite{lambin2012radiomics}. Through the extraction of handcrafted descriptors encoding shape, intensity distributions, and spatial texture patterns, radiomics aims to capture aspects of tissue heterogeneity that are not readily accessible through visual inspection alone. In recent years, this paradigm has become a natural testing ground for advanced machine‑learning strategies, particularly in settings characterized by limited sample sizes, heterogeneous acquisition protocols, and intrinsically overlapping clinical classes  \cite{STEFANO2024108827}. In this work, two well‑established radiomics benchmarks are considered, each representative of a distinct but complementary biomedical challenge. The first concerns the phenotyping of NSCLC\footnote{\textcolor{black}{NSCLC stands for non-small-cell lung carcinoma, the most common broad category of lung cancer, including several histopathological subtypes such as adenocarcinoma and squamous cell carcinoma.}} subtypes from multicenter CT \footnote{\textcolor{black}{CT stands for computed tomography, an imaging modality that uses X-rays to generate cross-sectional images of the body.}}  data  \cite{Pasini2023}, while the second addresses prostate cancer (PCa) risk stratification using $[^{18}\mathrm{F}]$PSMA-1007 PET/CT \footnote{\textcolor{black}{PET/CT stands for positron emission tomography/computed tomography, a hybrid imaging modality that combines metabolic information from PET with anatomical information from CT.}}  \cite{Pasini2025}. An example for the two imaging modalities (i.e., CT and PET/CT) and two considered case studies (NSCLC and PCa) is shown in Figure \ref{fig:example_NSCLC_PROSTATE}. Both tasks exemplify scenarios where radiomics pipelines have achieved competitive but not definitive performance, thus offering a meaningful context in which to evaluate whether a truly multiclass, quantum-mechanically inspired decision rule can provide a valid alternative.} A substantial body of recent work has demonstrated that both QML and quantum-inspired methodologies can be meaningfully stress-tested on real-world biomedical datasets, with medical imaging emerging as a particularly stringent domain because it combines high dimensionality, acquisition-driven variability, and clinically constrained sample sizes \cite{litjens2017survey, havlicek2019supervised, li2021quantum, quantum_bio_review, gupta2025systematic, Sergioli2021}.

\begin{figure}[h!]
    \includegraphics[width=1\linewidth]{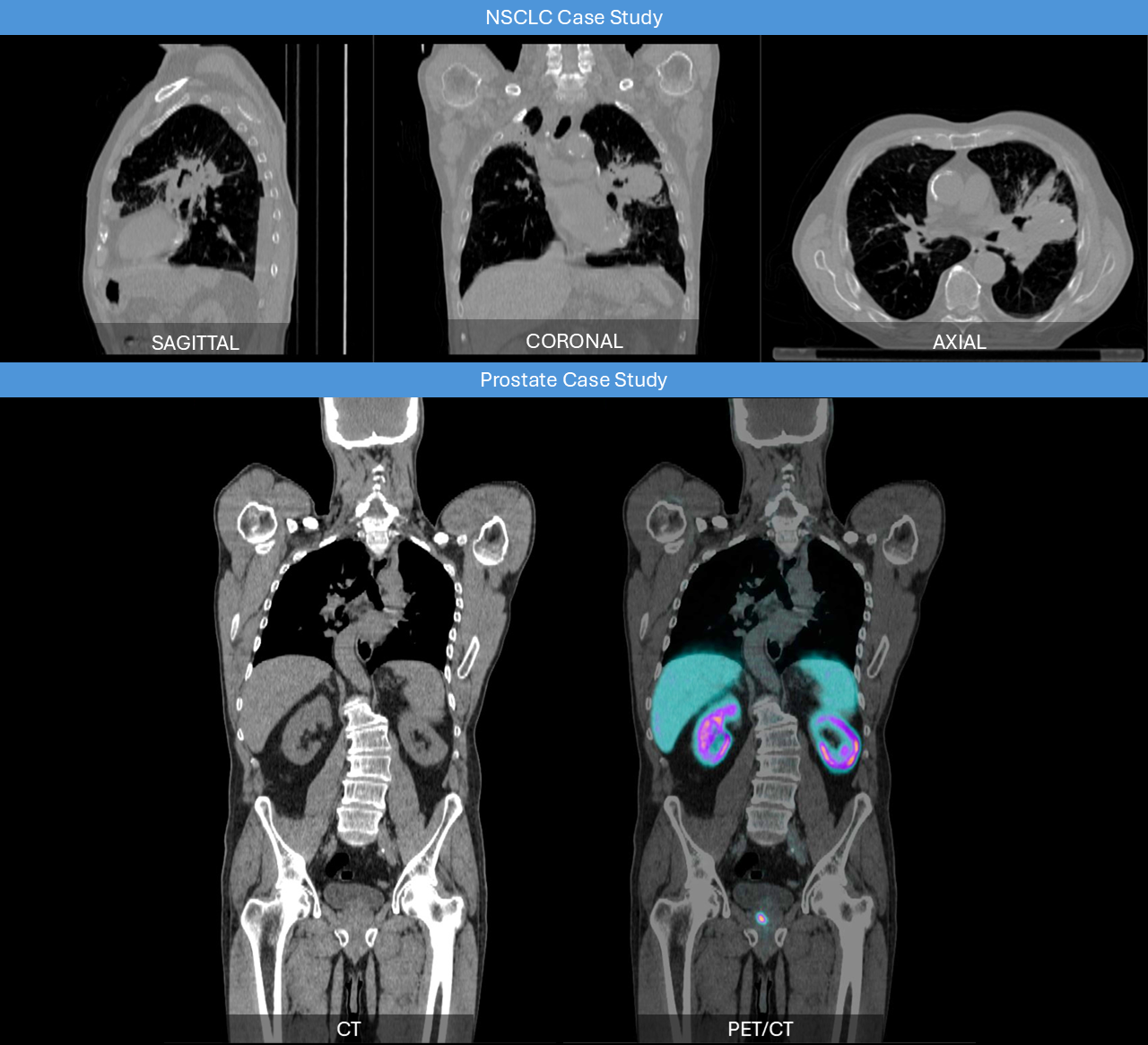}
    \caption{\textcolor{black}{Representative examples of the imaging modalities underlying the two benchmark datasets.} In the first row (NSCLC Case Study) there is an example of CT at the three anatomical planes: sagittal (left), coronal (middle), axial (right). In the second row, there is an example of CT (on the left) and PET/CT (on the right) for the PCa Case Study. PET/CT highlights high metabolic areas, in this case in the liver, kidneys, spleen and a focal spot in the prostate. \textcolor{black}{The figure is included for visual illustration only; the classification analysis is performed on radiomics features extracted from the segmented volumes of interest, as detailed in Section \ref{mm}.}}
    \label{fig:example_NSCLC_PROSTATE}
\end{figure}

{\textcolor{black}{In the NSCLC setting, radiomics is applied to distinguish among multiple histopathological subtypes, including Adenocarcinoma, Squamous Cell Carcinoma (SCC), Large Cell Carcinoma (LCC), and a residual “Not Otherwise Specified” (NOS) category. From a methodological standpoint, this problem is particularly demanding: class boundaries are radiologically subtle, label heterogeneity is substantial, and acquisition‑driven variability across centers often dominates phenotype‑related differences. Previous studies have shown that, even after standardized preprocessing and feature harmonization, multi‑class separation remains fragile, with performance degrading as the number of target classes increases. These characteristics make NSCLC subtype classification a stringent benchmark for evaluating classifiers that aim to operate directly in the multi‑class regime without resorting to binary decompositions. } Specifically, using two public databases, the authors extracted IBSI-compliant (Image Biomarker Standardisation Initiative) \cite{doi:10.1148/radiol.2020191145} radiomics features from manually and segmented lesions. The IBSI is an international effort aimed at establishing standardized definitions, nomenclature, and computational procedures for radiomics feature extraction. Its goal is to ensure that radiomics biomarkers are computed consistently and reproducibly across different software tools, imaging modalities, and research centers. By providing formal mathematical definitions, benchmark datasets, and reference values, IBSI helps improve the transparency, comparability, and clinical reliability of radiomics studies. A central focus of the work presented in  \cite{Pasini2023}  is the quantitative characterization of batch effects stemming from heterogeneous CT acquisition protocols, differences in scanner vendors, and variability in segmentation workflows. Through statistical analyses, the authors show that multicenter radiomics datasets tend to cluster primarily according to the acquisition site rather than the underlying tumor phenotype, even after applying standard preprocessing steps such as isotropic voxel resampling. Consequently, the ComBat harmonization technique  \cite{orlhac2022guide}   is employed to strike a balance between mitigating batch effects and preserving biologically meaningful discriminative information. Downstream learning is carried out using a repeated outer‑loop approach based on multiple stratified 80/20 train–test splits. Within each outer loop, feature selection is performed exclusively on the training folds through a two‑step procedure: an initial Kruskal–Wallis screening, followed by the Least Absolute Shrinkage and Selection Operator (LASSO) \cite{LASSO}. LASSO is among the most widely adopted radiomics feature‑reduction methods, as it enables the extraction of a compact set of stable and informative predictors while effectively limiting overfitting—a critical concern given the high feature‑to‑sample ratio that characterizes multicenter datasets. Model selection and hyperparameter optimization are handled by repeated 5-fold cross-validated Bayesian optimization. Six different classification models are compared using the harmonized datasets: Discriminant Analysis (DC), K-Nearest Neighbors (KNN), Support Vector Machines (SVM), Naïve Bayes (NB), Decision Trees (TREE), and Ensemble methods. 
{\textcolor{black}{Table~\ref{tab:classification_tasks} outcomes highlight the intrinsic difficulty of multi-class radiomics phenotyping.}

\begin{table}[htbp]
    \centering
    \caption{\textcolor{black}{Best reported classification accuracies from \cite{Pasini2023} across NSCLC radiomics tasks of increasing complexity.}}
    \label{tab:classification_tasks}
    \begin{tabular}{lll}
        \toprule
        \textbf{Classification task} & \textbf{Number of classes} & \textbf{Accuracy (Model)}\\
        \midrule
        SCC vs. ADC & 2 & 58.57\% (NB)\\
        \addlinespace
        SCC, ADC, LCC & 3 & 52.9\% (Ensemble)\\
        \addlinespace
        SCC, ADC, LCC, NOS & 4 & 56.2\% (Ensemble)\\
        \bottomrule
    \end{tabular}
    
\end{table}

In terms of the Area Under the Curve (AUC), the Ensemble model consistently yields the strongest results across every classification task. It achieves a notable AUC of $95.7\%$ for the ``Not Otherwise Specified'' class. For the other subtypes, AUC values ranged from $61.48\%$ to $79.9\%$ for Adenocarcinoma, $65.8\%$ to $69.1\%$ for Squamous Cell Carcinoma, and $70.6\%$ to $73.0\%$ for Large Cell Carcinoma. \textbf{
}
{\textcolor{black}{The prostate cancer (PCa) benchmark addresses a different, yet equally challenging, dimension of radiomics‑based learning \cite{Pasini2025}. Here, the objective is not histological phenotyping but clinical risk stratification, typically formulated as a binary discrimination between low‑ and high‑risk disease based on Gleason score (GS), a histological grading system (ranging from 1 to 5) assigned to the two most prevalent patterns in a biopsy sample  \cite{Ilick3519, AHMED2017815}. PET/CT imaging with PSMA‑targeted radiotracers enables a functional characterization of tumor biology beyond anatomical information alone,}  \textcolor{black}{but also introduces tracer‑specific interpretative pitfalls. In particular, while $[^{18}\mathrm{F}]$PSMA-1007 offers practical advantages in terms of half‑life and urinary excretion \cite{FGiesel}, it is known to exhibit unspecific bone uptake, especially in ribs and pelvic structures \cite{Bauckneht2025, Bauckneht2024, Laudicella2023}. In the considered benchmark, this limitation is mitigated by design: radiomics features are extracted exclusively from intraprostatic primary lesions, and clinical risk labels are derived solely from histopathological assessment, effectively decoupling the learning task from extra‑prostatic bone findings. Specifically, the PCa study  \cite{Pasini2025} addresses these clinical needs by focusing on binary risk stratification in a retrospective cohort where tumor regions are manually delineated and radiomics features are extracted following Standardized Uptake Value (SUV) normalization, as reported in \cite{https://doi.org/10.1002/ima.22154}.} } The SUV is the most common semi-quantitative parameter used to estimate biodistribution in PET images. The SUV normalizes the voxel activity considering acquisition time, administered activity, and patient weight. In this way, PET images took into account factors that would otherwise be ignored during radiomics analysis.  The key contribution of  \cite{Pasini2025}   lies in the explicit operationalization of robustness with respect to dataset variability. A preliminary pipeline—iterated 30 times and comprising an 80/20 stratified split, LASSO‑based feature selection, and five‑fold cross‑validation for Bayesian hyperparameter optimization—serves to identify stable features and to define a pool of candidate base learners. Subsequently, an ensemble model is constructed and evaluated across the same repeated splits. Trained on a feature set that combines highly frequent “robust” predictors with a decorrelated subset of moderately frequent “fine‑tuning” features, the ensemble achieves the strongest average test performance: an accuracy of 79.52\% and an AUC of 85.75\%. 

Taken together, these two baselines illustrate the potential of state‑of‑the‑art radiomics: they show that competitive performance is attainable through rigorous validation, stability‑aware feature selection, and explicit treatment of acquisition‑driven heterogeneity. At the same time, they expose the bottlenecks that motivate the present work—most notably, the fragility of multi‑class structures in high‑dimensional representation spaces. These characteristics render the datasets ideal benchmark testbeds for evaluating whether quantum‑inspired multi‑class decision rules can provide a more principled inductive bias under realistic clinical constraints.

\section{Materials and Methods}\label{mm}

In this section we present the results of a direct comparison between the experiments described in the previous section and the PGM-based approach. The PGM classifier is evaluated on the same datasets introduced above, using a matched experimental setting, and its performance is reported side by side with the baseline results.

\subsection{Dataset Description and Image Processing}

This section describes the datasets and the processing procedures adopted to benchmark the proposed quantum‑inspired classifier. The analysis focuses on the two clinical scenarios introduced earlier: phenotyping of NSCLC and risk stratification in PCa. Starting from CT images in the first scenario and PET/CT images in the second, the overarching aim was to quantitatively characterize the target lesions through a standardized radiomics workflow built upon the \textit{matRadiomics} framework \cite{jimaging8080221}.
\textit{matRadiomics} is a comprehensive, IBSI‑compliant radiomics platform designed to support the entire analytical pipeline — from image visualization, to the segmentation process, to image pre-processing, to feature extraction, to data harmonization and predictive modeling. Built on the PyRadiomics \cite{compRadiomics}, it enables the standardized computation of 1,781 high‑dimensional features, including shape descriptors, first‑order statistics, and a broad set of texture features derived from the Gray Level Co‑occurrence Matrix, Gray Level Run Length Matrix, Gray Level Size Zone Matrix, Neighboring Gray Tone Difference Matrix, and Gray Level Dependence Matrix.
For both CT and PET modalities, features were then extracted from three complementary image representations:
\begin{enumerate}
    \item Original biomedical images,
    \item Laplacian of Gaussian (LoG) filtered images, and
    \item Multilevel wavelet‑decomposed images.
\end{enumerate}
\textit{matRadiomics} ensures high reproducibility, transparent control of preprocessing parameters, and robust handling of multicenter data through built‑in batch‑effect detection and correction tools such as ComBat harmonization, an essential component when working with heterogeneous acquisition protocols \cite{STEFANO2024108827}.
The specific configurations for the two studies are detailed in the following subsections.

\subsubsection{Non-Small-Cell Lung Carcinoma (NSCLC) Dataset}
\label{subsubsec:NSCLC_dataset_details}
The lung cancer cohort was established by merging two public repositories: \textcolor{black}{the NSCLC-Radiomics \cite{Aerts2014} and NSCLC-Radiogenomics \cite{Bakr2018} datasets}, resulting in 466 patients categorized into four histopathological subtypes: Squamous Cell Carcinoma (SCC, $N=152$), Adenocarcinoma (ADC, $N=150$), Large Cell Carcinoma (LCC, $N=106$), and Not Otherwise Specified (NOS, $N=58$). \textcolor{black}{Here, $N$ denotes the number of patients in the corresponding subgroup.} To evaluate the proposed classifier across different diagnostic complexities, the original cohort was partitioned into three distinct datasets with varying labeling schemes:
\begin{itemize}
    \item A \textbf{4-classes dataset} (4-c) comprising all histopathological subtypes (SCC, ADC, LCC, and NOS; $N=466$).
    \item A \textbf{3-classes dataset} (3-c) excluding the NOS subtype ($N=408$).
    \item A \textbf{2-classes dataset} (2-c) focusing on the most prevalent subtypes, Squamous Cell Carcinoma and Adenocarcinoma ($N=302$).
\end{itemize}
To mitigate batch effects inherent in multicenter CT data acquired from different manufacturers (Siemens, CMS, GE, Philips), images were resampled to an isotropic voxel size of $1 \times 1 \times 1 \text{ mm}^3$ using linear interpolation \cite{Pasini2023}. The volumes of interest (VOIs) to be examined were provided together with the CT images. Feature harmonization was subsequently performed via the ComBat tool to standardize the mean and variance across the different acquisition centers. \textcolor{black}{ Description of Datasets, Institutions, Manufacturers, Geometric Information and Acquisition Protocol is summarized in Table \ref{tab:nsclc_scanners}}.
\textcolor{black}{Regarding feature extraction, grey-level discretization was performed using a fixed bin count of 64, wavelet decompositions were performed utilizing the Haar transform and LoG filtering utilizing a sigma value in a range from 0.5 to 5.0, with a step of 0.5}
\textcolor{black}{\begin{table}[htbp]
    \centering
    \caption{\textcolor{black}{Per-scanner and protocol details for the Non-Small-Cell Lung Carcinoma (NSCLC) cohort.}}
    \label{tab:nsclc_scanners}
    \resizebox{\textwidth}{!}{
    \begin{tabular}{p{4.5cm}lp{3.5cm}llp{1.5cm}p{4.5cm}}
        \toprule
        \textbf{Dataset} \newline \textbf{(Institution)} & \textbf{Modality} & \textbf{Manufacturer \& Scanner} & \textbf{Pixel Spacing (x, y) [mm]} & \textbf{Slice Thickness (z) [mm]} & \textbf{Matrix} & \textbf{Acquisition Parameters} \\
        \midrule
        NSCLC-Radiomics \newline (MAASTRO Clinic) & CT & Siemens & 0.9765625, 0.9765625 & 3.0 & 512 $\times$ 512 & NA \\
        \addlinespace
        NSCLC-Radiomics \newline (MAASTRO Clinic) & CT & CMS & 0.9770, 0.9770 & 3.0 & 512 $\times$ 512 & NA\\
        \addlinespace
        NSCLC-Radiogenomics \newline (Stanford \& Palo Alto VA) & CT & Siemens, GE Medical Systems, Philips & Range: 0.589844 -- 0.976562 & Range: 0.625 -- 3.0 (median 1.5) & 512 $\times$ 512 & Tube Voltage: 80--140 kVp (mean 120)\newline Tube Current: 124--699 mA (mean 220) \\
        \bottomrule
        \multicolumn{7}{l}{\footnotesize} \\
    \end{tabular}
    }
\end{table}}

\subsubsection{Prostate Cancer (PCa) Dataset}

The dataset included patients who underwent \textcolor{black}{$[^{18}\mathrm{F}]$PSMA-1007} PET/CT imaging for PCa staging. All subjects were retrospectively selected according to predefined inclusion criteria, namely: histologically confirmed PCa, availability of Gleason grade, and a positive PET scan. \textcolor{black}{To ensure patients were strictly treatment-naïve at the time of imaging, patients previously treated with surgery, radiotherapy, chemotherapy, endocrine therapy, or any other intervention prior to the PET scan, as well as those with a history of other malignancies, were excluded. The final cohort comprised 143 patients. To mitigate potential sampling bias associated with biopsy-derived scores, the cohort was stratified according to their definitive post-surgical Gleason grades} (72 with high-risk and 71 with low-risk). PET/CT acquisitions were acquired approximately 120 minutes after tracer injection. Manual slice-by-slice segmentation was performed by experienced clinicians to define the VOIs. \textcolor{black}{The specific scanner hardware, image resolution metrics, and acquisition protocols utilized for the cohort are summarized in Table \ref{tab:pca_scanners}. 
To ensure quantitative accuracy and reproducibility for the subsequent radiomics feature extraction, PET image reconstruction and correction methods were standardized according to the respective scanner hardware. For acquisitions on the General Electric Discovery system, PET images were reconstructed utilizing the proprietary VPFX algorithm, which corresponds to a 3D Ordered Subsets Expectation Maximization (OSEM) algorithm incorporating Time-of-Flight (TOF) and Point Spread Function (PSF) modeling. Comprehensive standard corrections were applied, including radioactive decay, attenuation, dead time, detector calibration, slice sensitivity, and normalization. Scatter correction was applied using a model-based approach, and random coincidences were corrected utilizing a singles-based estimation method (SING). Conversely, for the Siemens Biograph system, reconstruction was also performed utilizing a 3D OSEM algorithm with PSF and TOF modeling, specifically configured with 4 iterations and 10 subsets (denoted as PSF+TOF 4i10s). Attenuation correction was derived from measured whole-body CT data (AC CT WB). Scatter correction was model-based with relative scatter scaling, and randoms were corrected using the delayed coincidence window method (DLYD).}
\textcolor{black}{Regarding feature extraction, voxel intensities were first standardized to the body weight-based Standardized Uptake Value (SUVbw) measured in g/ml, with no explicit partial-volume correction (PVC) applied. Images were then resampled to an isotropic voxel resolution of $2 \times 2 \times 2 \text{ mm}^3$ utilizing B-spline interpolation. Image normalization was enabled (scale=1), and grey-level discretization was executed using a fixed bin width of 0.25. Wavelet decompositions were performed utilizing the Coiflet 1 basis function. Log filtering was based on sigma values in a range between 0.5 and 5, with a step of 0.5.} \textcolor{black}{A structured side-by-side summary of the feature extraction parameters adopted in the two case studies is provided in Table~\ref{tab:appendix_extraction} of Appendix~\ref{app:extraction_params}.}

\begin{table}[htbp]
    \centering
    \caption{\textcolor{black}{Per-scanner and protocol details for the Prostate Cancer (PCa) cohort.}}
    \label{tab:pca_scanners}
    \resizebox{\textwidth}{!}{
    \begin{tabular}{llcllp{5cm}}
        \toprule
        \textbf{Scanner Manufacturer \& Model} & \textbf{Modality} & \textbf{Patients ($n$)} & \textbf{Matrix Size} & \textbf{Voxel Size (x, y, z) [mm$^3$]} & \textbf{Specific Acquisition Parameters} \\
        \midrule
        General Electric & CT (16 row helical) & 68 & 512 $\times$ 512 & 1.37 $\times$ 1.37 $\times$ 3.75 & Tube Voltage: 140 kVp\newline Tube Current: 800 mAmax \\
        \cmidrule{2-6}
        Discovery 690FX \& MOT & PET ($^{18}$F-PSMA-1007) & & 256 $\times$ 256 & 2.73 $\times$ 2.73 $\times$ 3.27 & Acq. Time: 90 s/bed (7--8 beds)\newline Uptake Time: $\sim$120 min\newline Dose: $\sim$4 MBq/kg \\
        \midrule
        Siemens & CT (16 row helical) & 75 & 512 $\times$ 512 & 0.98 $\times$ 0.98 $\times$ 3.0 & Tube Voltage: 130 kVp\newline Tube Current: 345 mAmax \\
        \cmidrule{2-6}
        Biograph Horizon 4R & PET ($^{18}$F-PSMA-1007) & & 512 $\times$ 512 & 1.45 $\times$ 1.45 $\times$ 3.0 & Acq. Time: 90 s/bed (6--7 beds)\newline Uptake Time: $\sim$120 min\newline Dose: $\sim$4 MBq/kg \\
        \bottomrule
    \end{tabular}
    }
\end{table}

\subsection{Experimental Protocol}
\label{sec:ex_protoc}
To ensure the comparability of our results with existing benchmarks, we adopted the data partitioning and pre-processing frameworks established in previous studies by Pasini \textit{et al.}~\cite{Pasini2023, Pasini2025}. \\

For the NSCLC study, our experimentation utilized the identical feature-selected and harmonized data splits from the original study (comprising 10 repetitions). This framework was applied to three distinct classification tasks with varying histopathological granularity---namely, \textbf{4-class} (SCC, ADC, LCC, NOS), \textbf{3-class} (SCC, ADC, LCC), and \textbf{binary} (SCC vs.~ADC) datasets---as detailed in Subsection~\ref{subsubsec:NSCLC_dataset_details}. For each classification task, we implemented an optimization protocol consisting of 10 repetitions of the grid search with stratified 5-fold cross-validation on the training set to determine the optimal hyperparameters (optimizing for AUC), using the parameter space detailed in Table~\ref{tab:grid_space}. The optimized model was then evaluated on the independent test set. To ensure statistical stability, performance metrics were first averaged across the ten iterations of the 5-fold cross-validation within each split, and then the final results were obtained by averaging these values across all ten splits.\\

For the prostate cancer dataset, the experimental protocol was designed to ensure full consistency with the benchmarking ensemble model by adopting the exact 30 stratified train/test splits (80/20 ratio) established in the reference study by Pasini~\textit{et al.}~\cite{Pasini2025}.
This approach enabled the use of pre-calculated feature frequency data from 30 iterations of the LASSO selection process, replicating the 13 feature subsets employed in the original experimentation.

These subsets include eleven frequency-based groups (generated using thresholds from $\geq 90\%$ down to $0\%$, including an additional $5\%$ threshold) and two specialized subsets: the \textit{finetuning} subset and the \textit{features7030r} subset (a union of \textit{finetuning} and \textit{features70}). \textcolor{black}{For consistency with the benchmark, the composition of these subsets was determined according to the procedure originally established in~\cite{Pasini2025}: the eleven frequency-based groups follow directly from the selection-frequency thresholds applied to the 30 LASSO iterations of the reference pipeline, while the \textit{finetuning} subset retains those features with a selection frequency between $30\%$ (included) and $70\%$ (not included) whose Pearson Correlation Coefficient (PCC) with respect to the features in \textit{features70} is below $0.3$, with the PCC value averaged over the 30 iterations. A complete and self-contained description of all 13 feature subsets, including their construction criteria, sizes, the explicit composition of both the \textit{features70} and \textit{finetuning} subsets together with the corresponding average PCC values, and the selection frequency of all 79 features retained at least once across the 30 iterations (from which the composition of any frequency-based subset can be reconstructed), is provided for the reader's convenience in Tables~\ref{tab:appendix_subsets}, \ref{tab:appendix_pcc} and \ref{tab:appendix_all_features} of Appendix~\ref{app:feature_subsets}.}

Hyperparameter optimization was performed using a grid search approach with stratified 5-fold cross-validation on the training sets, with the parameter space defined in Table~\ref{tab:grid_space}, using the Area Under the Curve (AUC) as the objective metric.
The most robust configuration was selected based on two criteria: (1) the frequency of occurrence across the 30 independent runs, and (2) in cases of ties, the mean test-set AUC achieved by each candidate configuration.
This two-step selection process ensures both stability across diverse data partitions and optimal predictive performance, mitigating bias toward any specific data split. According to this criterion, the most frequently selected configuration employed amplitude-based encoding, no tensor copies of the feature vectors, and a rescaling factor of 0.5.

The classification performance on the test set for both datasets was evaluated using a comprehensive set of metrics. For the Lung Carcinoma study, we considered Accuracy and AUC per-class, while for the PCa study, the evaluation framework included  Accuracy, AUC, Precision, Recall (Sensitivity), F1-score, and Specificity, ensuring a rigorous assessment of the model's ability to distinguish between high-risk and low-risk patients.

\begin{table}[htbp]
    \centering    
    \caption{\textcolor{black}{Hyperparameter space employed for grid search in the NSCLC (2-class, 3-class, and 4-class) and PCa experiments.}}
    \label{tab:grid_space}
    \begin{tabular}{lll}
        \toprule
        \textbf{Encoding} & \textbf{Rescaling} & \textbf{Number of Copies}\\ 
        \midrule
        \addlinespace
        \{stereo, amplit\} & \{0.5, 1, 2, 4, 8, 16\} & \{1, 5, 10, 15, 20, 25, 30, 35, 40, 45, 50, 55, 60\}\\
        \bottomrule
    \end{tabular}
\end{table}

\newpage
\section{Results}

In this section we report the outcomes of the experiments conducted on the two
benchmark datasets (NSCLC and PCa).  For each dataset we present the
evaluation metrics that enable a direct, quantitative comparison between our
PGM classifier and the results reported in ~\cite{Pasini2023, Pasini2025}.

Overall, the presented results demonstrate that our PGM approach attains
competitive (and in several cases superior) performance relative to the
state‑of‑the‑art methods evaluated in the literature.

\subsection{NSCLC Dataset}

For the NSCLC dataset we provide three Figures (\ref{fig:NSCLC-metrics-2},
\ref{fig:NSCLC-metrics-3}, and~\ref{fig:NSCLC-metrics-4}), one for each class
configuration (2, 3, and 4 classes).  Each figure reports the mean AUC computed over
all data splits together with macro-averaged accuracy score for the PGM classifier and the classical machine learning models exploited by Pasini et al. \cite{Pasini2023}. 
For each labeling we also present a win‑loss heatmap \ref{fig:NSCLC-win-loss} that juxtaposes the conventional classifiers with the PGM classifier. The heatmap is generated by averaging the per‑class AUC scores of each model across cross‑validation folds, thereby illustrating the overall performance trend. Because only aggregate values are available for the baseline classifiers, statistical significance could not be evaluated. \textcolor{black}{The mean confusion matrices across the 10 splits for the 2-class, 3-class, and 4-class configurations are reported in Figure~\ref{fig:cm_nsclc}, providing a per-class breakdown of the classification performance across the NSCLC histopathological subtypes. Detailed per-class performance metrics with 95\% confidence intervals are reported in Tables~\ref{tab:pgm_2class_adc}, \ref{tab:pgm_3class}, and~\ref{tab:pgm_4class} for the 2-class, 3-class, and 4-class configurations, respectively.}

\begin{figure*}[h!]
    \includegraphics[width=1\linewidth]{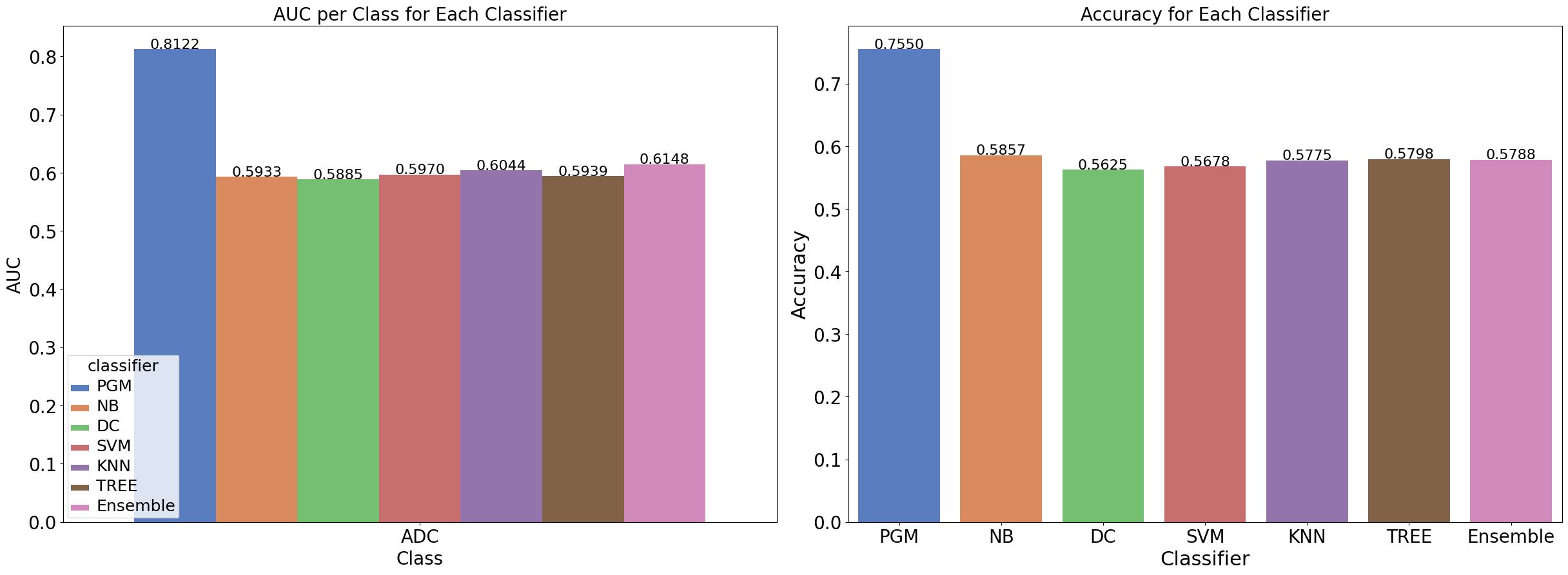}
    \caption{ Comparative evaluation of the proposed PGM classifier against conventional baseline models on the NSCLC two-class dataset. (a) Mean test area under the ROC curve (AUC) reported for each class. (b) Macro-averaged accuracy across the two classes.}
    \label{fig:NSCLC-metrics-2}
\end{figure*}

\begin{figure*}[h!]
    \includegraphics[width=1\linewidth]{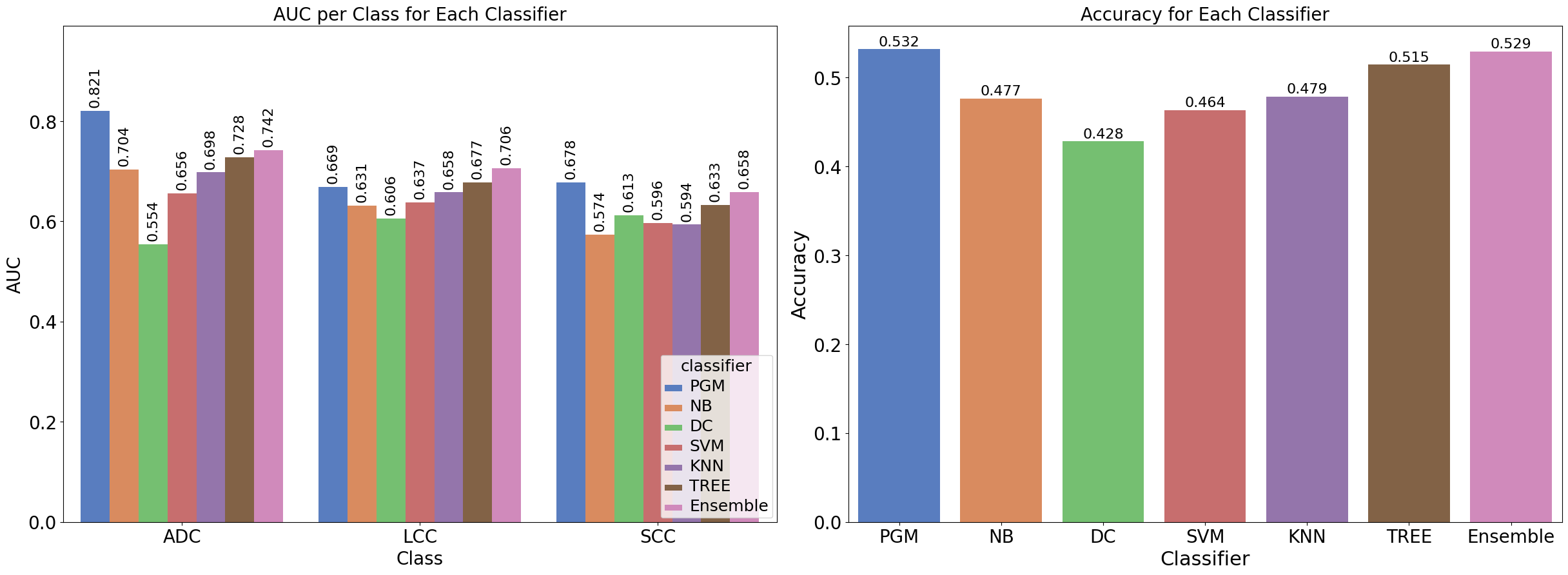}
    \caption{Comparative evaluation of the proposed PGM classifier against conventional baseline models on the NSCLC three-class dataset. (a) Mean test area under the ROC curve (AUC) reported for each class. (b) Macro-averaged accuracy across the three classes.}
    \label{fig:NSCLC-metrics-3}
\end{figure*}

\begin{figure*}[h!]
    \includegraphics[width=1\linewidth]{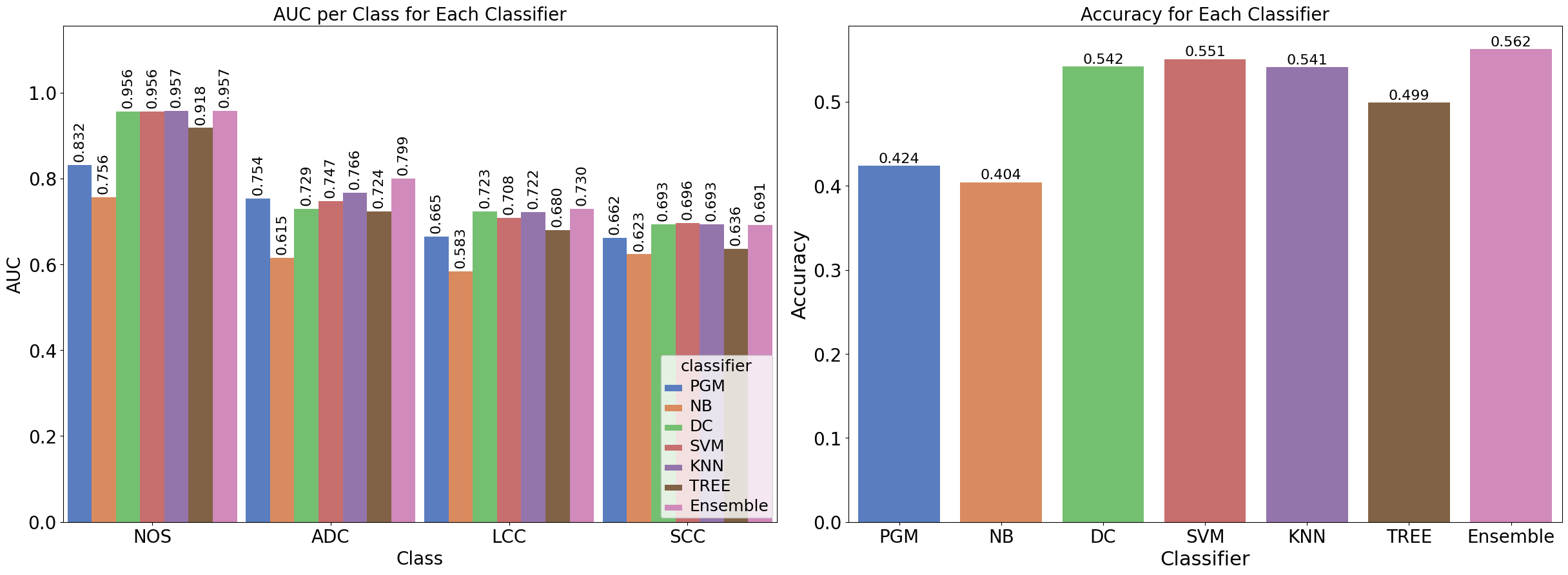}
    \caption{Comparative evaluation of the proposed PGM classifier against conventional baseline models on the NSCLC four-class dataset. (a) Mean test area under the ROC curve (AUC) reported for each class. (b) Macro-averaged accuracy across the four classes.}
    \label{fig:NSCLC-metrics-4}
\end{figure*}

%__________________________________________________________

\begin{figure}[h!]
    \includegraphics[width=1\linewidth]{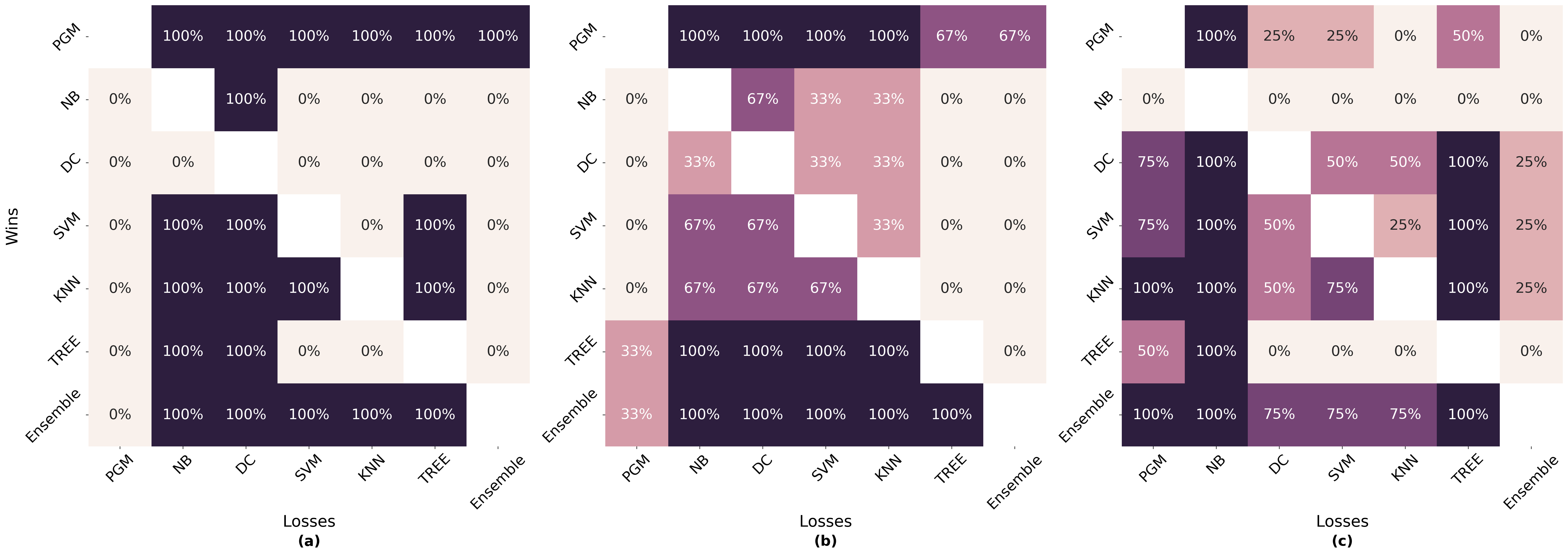}
    \caption{Win–loss heatmap comparing classifiers for each labeling scenario. The three labeling scenarios are:
        \textbf{(a)} 2‑class labeling – only two disease categories are considered;
        \textbf{(b)} 3‑class labeling – three distinct categories are distinguished;
        \textbf{(c)} 4‑class labeling – all four histopathological subtypes are used.
        For each pair of classifiers, each cell reports the proportion of classes for which the
        row classifier attains a higher AUC than the column classifier.  Values therefore range
        from 0$\%$ to 100$\%$ and indicate the fraction of class‑wise pairwise wins.  \textcolor{black}{Darker cells}
        denote a larger proportion of class‑level dominance.}
    \label{fig:NSCLC-win-loss}
\end{figure}

%-------------------------- TAB METRICHE e CI 
\begin{table}[h!]
\centering
\caption{Performance of the PGM classifier on the NSCLC 2-class dataset (ADC vs.\ SCC),
         expressed as mean\,$\pm$\,semi-amplitude of the 95\% confidence interval over 10 train/test splits. Results are reported for the adenocarcinoma (ADC) class.}
\label{tab:pgm_2class_adc}
\resizebox{\textwidth}{!}{%
\begin{tabular}{lcccccc}
\toprule
 & \textbf{Accuracy} & \textbf{AUC} & \textbf{Sensitivity} & \textbf{Specificity} & \textbf{Precision} & \textbf{F1-score} \\
\midrule
Validation & $0.7430 \pm 0.0301$ & $0.7997 \pm 0.0266$ & $0.7242 \pm 0.1005$ & $0.7615 \pm 0.0419$ & $0.7515 \pm 0.0122$ & $0.7294 \pm 0.0563$ \\
Test       & $0.7550 \pm 0.0413$ & $0.8122 \pm 0.0330$ & $0.7333 \pm 0.1018$ & $0.7767 \pm 0.0738$ & $0.7746 \pm 0.0509$ & $0.7434 \pm 0.0624$ \\
\bottomrule
\end{tabular}%
}
\end{table}

% ── 3-class table ────────────────────────────────────────────────────────────
\begin{table}[h!]
\centering
\caption{Performance of the PGM classifier on the NSCLC 3-class dataset
         (ADC, LCC, SCC), expressed as mean\,$\pm$\,semi-amplitude of the 95\%
         confidence interval over 10
         train/test splits.}
\label{tab:pgm_3class}
\resizebox{\textwidth}{!}{%
\begin{tabular}{llcccccc}
\toprule
\textbf{Class} & & \textbf{Accuracy} & \textbf{AUC} & \textbf{Sensitivity} & \textbf{Specificity} & \textbf{Precision} & \textbf{F1-score} \\
\midrule
\multirow{2}{*}{ADC}
  & Validation & $0.5300 \pm 0.0455$ & $0.7978 \pm 0.0374$ & $0.6733 \pm 0.1117$ & $0.7899 \pm 0.0672$ & $0.6600 \pm 0.0476$ & $0.6541 \pm 0.0655$ \\
  & Test       & $0.5321 \pm 0.0354$ & $0.8205 \pm 0.0205$ & $0.7267 \pm 0.0932$ & $0.7922 \pm 0.0780$ & $0.6952 \pm 0.0756$ & $0.6965 \pm 0.0347$ \\
\midrule
\multirow{2}{*}{LCC}
  & Validation & $0.5300 \pm 0.0455$ & $0.6638 \pm 0.0415$ & $0.5318 \pm 0.1597$ & $0.6855 \pm 0.1220$ & $0.3766 \pm 0.0379$ & $0.4208 \pm 0.0810$ \\
  & Test       & $0.5321 \pm 0.0354$ & $0.6694 \pm 0.0414$ & $0.5000 \pm 0.1297$ & $0.7033 \pm 0.1234$ & $0.3930 \pm 0.0466$ & $0.4156 \pm 0.0589$ \\
\midrule
\multirow{2}{*}{SCC}
  & Validation & $0.5300 \pm 0.0455$ & $0.6764 \pm 0.0374$ & $0.3877 \pm 0.1098$ & $0.8337 \pm 0.0529$ & $0.5589 \pm 0.0731$ & $0.4446 \pm 0.1089$ \\
  & Test       & $0.5321 \pm 0.0354$ & $0.6779 \pm 0.0401$ & $0.3600 \pm 0.1225$ & $0.8137 \pm 0.0632$ & $0.5175 \pm 0.0886$ & $0.4091 \pm 0.1152$ \\
\bottomrule
\end{tabular}%
}
\end{table}

% ── 4-class table ────────────────────────────────────────────────────────────
\begin{table}[h!]
\centering
\caption{Performance of the PGM classifier on the NSCLC 4-class dataset
         (NOS, ADC, LCC, SCC), expressed as mean\,$\pm$\,semi-amplitude of the
         95\% confidence interval over
         10 train/test splits.}
\label{tab:pgm_4class}
\resizebox{\textwidth}{!}{%
\begin{tabular}{llcccccc}
\toprule
\textbf{Class} & & \textbf{Accuracy} & \textbf{AUC} & \textbf{Sensitivity} & \textbf{Specificity} & \textbf{Precision} & \textbf{F1-score} \\
\midrule
\multirow{2}{*}{NOS}
  & Validation & $0.4501 \pm 0.0751$ & $0.8350 \pm 0.0703$ & $0.7537 \pm 0.0847$ & $0.7971 \pm 0.1360$ & $0.4071 \pm 0.0945$ & $0.5165 \pm 0.0993$ \\
  & Test       & $0.4237 \pm 0.0862$ & $0.8320 \pm 0.0918$ & $0.7689 \pm 0.1177$ & $0.7761 \pm 0.1525$ & $0.4009 \pm 0.1301$ & $0.5141 \pm 0.1433$ \\
\midrule
\multirow{2}{*}{ADC}
  & Validation & $0.4501 \pm 0.0751$ & $0.7357 \pm 0.0585$ & $0.5152 \pm 0.1421$ & $0.8413 \pm 0.0573$ & $0.5779 \pm 0.1257$ & $0.5326 \pm 0.1327$ \\
  & Test       & $0.4237 \pm 0.0862$ & $0.7537 \pm 0.0671$ & $0.4933 \pm 0.1850$ & $0.8295 \pm 0.0863$ & $0.5701 \pm 0.1884$ & $0.4976 \pm 0.1541$ \\
\midrule
\multirow{2}{*}{LCC}
  & Validation & $0.4501 \pm 0.0751$ & $0.6529 \pm 0.0348$ & $0.4187 \pm 0.0887$ & $0.7772 \pm 0.0448$ & $0.3534 \pm 0.0326$ & $0.3770 \pm 0.0553$ \\
  & Test       & $0.4237 \pm 0.0862$ & $0.6647 \pm 0.0412$ & $0.3857 \pm 0.0988$ & $0.7671 \pm 0.0830$ & $0.3440 \pm 0.0512$ & $0.3481 \pm 0.0415$ \\
\midrule
\multirow{2}{*}{SCC}
  & Validation & $0.4501 \pm 0.0751$ & $0.6654 \pm 0.0514$ & $0.2914 \pm 0.0650$ & $0.8627 \pm 0.0249$ & $0.4981 \pm 0.0501$ & $0.3629 \pm 0.0657$ \\
  & Test       & $0.4237 \pm 0.0862$ & $0.6617 \pm 0.0408$ & $0.2519 \pm 0.0758$ & $0.8742 \pm 0.0455$ & $0.5275 \pm 0.1655$ & $0.3235 \pm 0.0868$ \\
\bottomrule
\end{tabular}%
}
\end{table}

% ── Figura Confusion matrices NSCLC ───────────────────────────────────────
\begin{figure}[h!]
    \centering
    \includegraphics[width=\textwidth]{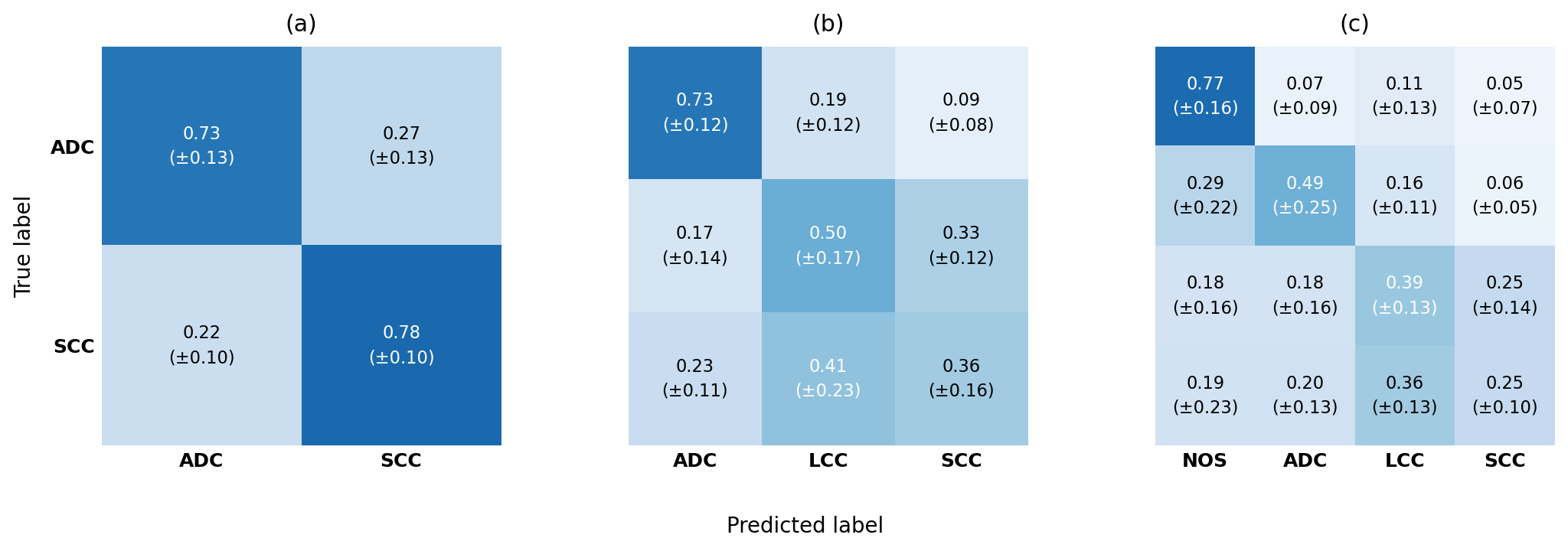}
    \caption{Mean confusion matrices (normalized by row) for the NSCLC
    histopathological subtype classification task, averaged over 10
    train/test splits. Each cell reports the mean value and the standard
    deviation (±std). Panels show the binary classification problem
    (a) SCC vs.\ ADC (2-class), (b) the 3-class problem (ADC, LCC, SCC),
    and (c) the 4-class problem (NOS, ADC, LCC, SCC).
    ADC: adenocarcinoma; LCC: large cell carcinoma;
    SCC: squamous cell carcinoma; NOS: not otherwise specified.}
    \label{fig:cm_nsclc}
\end{figure}

%\clearpage
\newpage
\subsection{PCa Dataset}

\begin{figure*}[h!]
    \centering
    
    \begin{subfigure}[t]{0.48\linewidth}
        \centering
        \includegraphics[width=\linewidth]{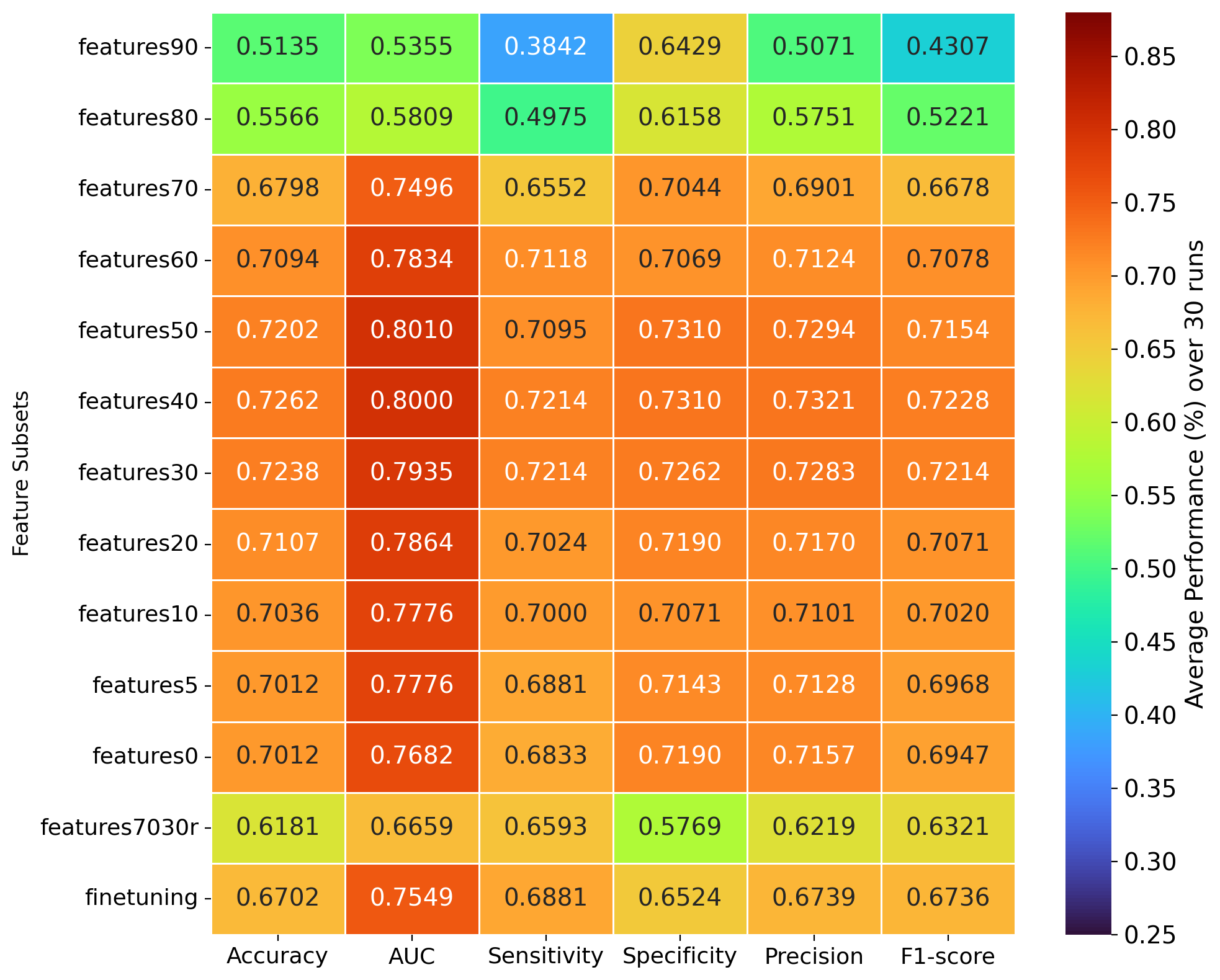}
        \caption{Mean performance of the PGM classifier across feature-set variations, averaged over 30 iterations using the same test-set subdivision of the original study.}
        \label{fig:PCa-mean-performance}
    \end{subfigure}
    \hfill
    \begin{subfigure}[t]{0.48\linewidth}
        \centering
        \includegraphics[width=1.1\linewidth]{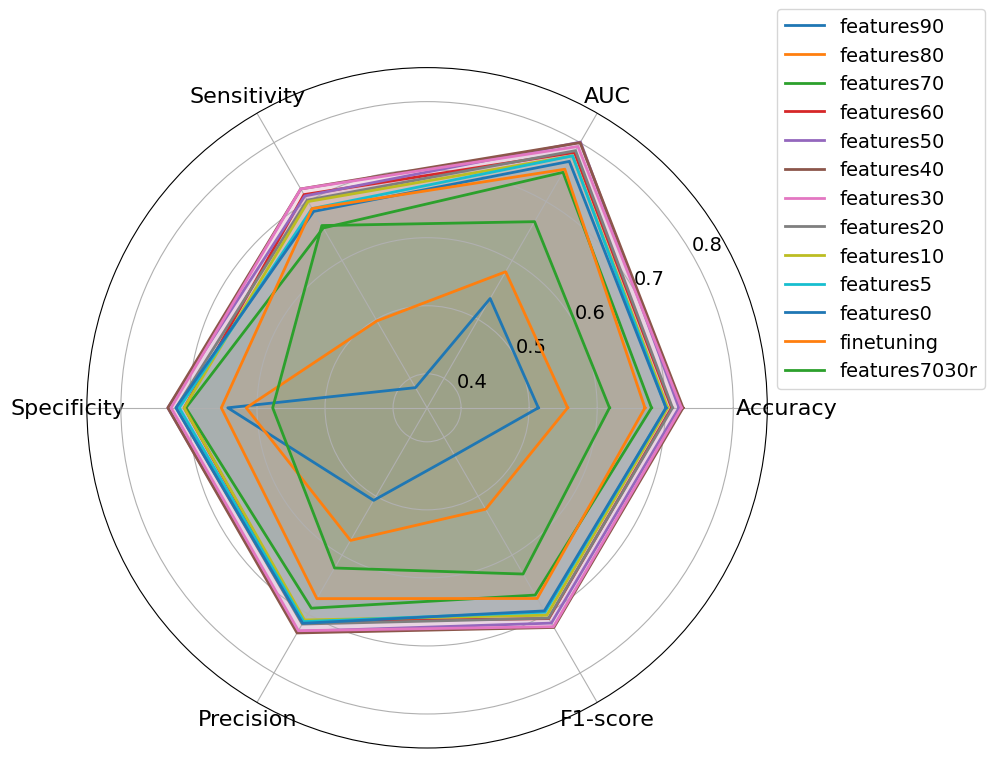}
        \caption{Radar chart summarizing the six evaluation metrics across feature-selection scenarios.}
        \label{fig:Radar}
    \end{subfigure}
    
    \caption{Performance analysis of the PGM classifier on the PCa dataset under different feature-selection scenarios.}
    \label{fig:PCa-combined}
\end{figure*}

For the PCa dataset, Figure \ref{fig:PCa-mean-performance} shows the mean
performance of the PGM classifier across the different feature‑set
variations.  This plot allows a straightforward comparison with the
quantum‑inspired classifier and with the ensemble model results presented in
Figure~6 of Pasini et al. \cite{Pasini2025}.
Complementarily, Figure~\ref{fig:Radar} presents a radar chart summarizing the six evaluation metrics across feature-selection scenarios. This representation provides an integrated view of the multidimensional performance profile of the PGM classifier and facilitates an immediate qualitative comparison of the balance among the considered metrics.
To highlight the relative behaviour of the two approaches across the 13 predefined feature subsets, we compute the element‑wise difference of the six performance metrics (Accuracy, AUC, Sensitivity, Specificity, Precision, F1‑score) between the PGM and the Ensemble models.  The resulting matrix is visualised in Figure~\ref{fig:pgm‑ensemble‑diff‑heatmap}.  The heat‑map employs a diverging blue–red colour map centred at zero: blue cells indicate a positive difference (PGM outperforms the Ensemble), while red cells denote a negative difference (Ensemble outperforms PGM).  The intensity of the colour is proportional to the magnitude of the difference, allowing a rapid visual assessment of the scenarios in which the quantum‑inspired classifier provides a tangible advantage over the traditional ensemble method. \textcolor{black}{The mean confusion matrices across the 30 splits for all 13 feature subsets are reported in Figure~\ref{fig:cm_pca}, providing a per-class breakdown of the classification performance between high-risk and low-risk patients. A detailed numerical summary of the PGM classifier performance across all 13 feature subsets, including 95\% bootstrap confidence intervals for each metric, is provided in Table~\ref{tab:pgm_pca_subsets}. }

\begin{figure}[htbp]
    \centering
    \includegraphics[width=0.9\linewidth]{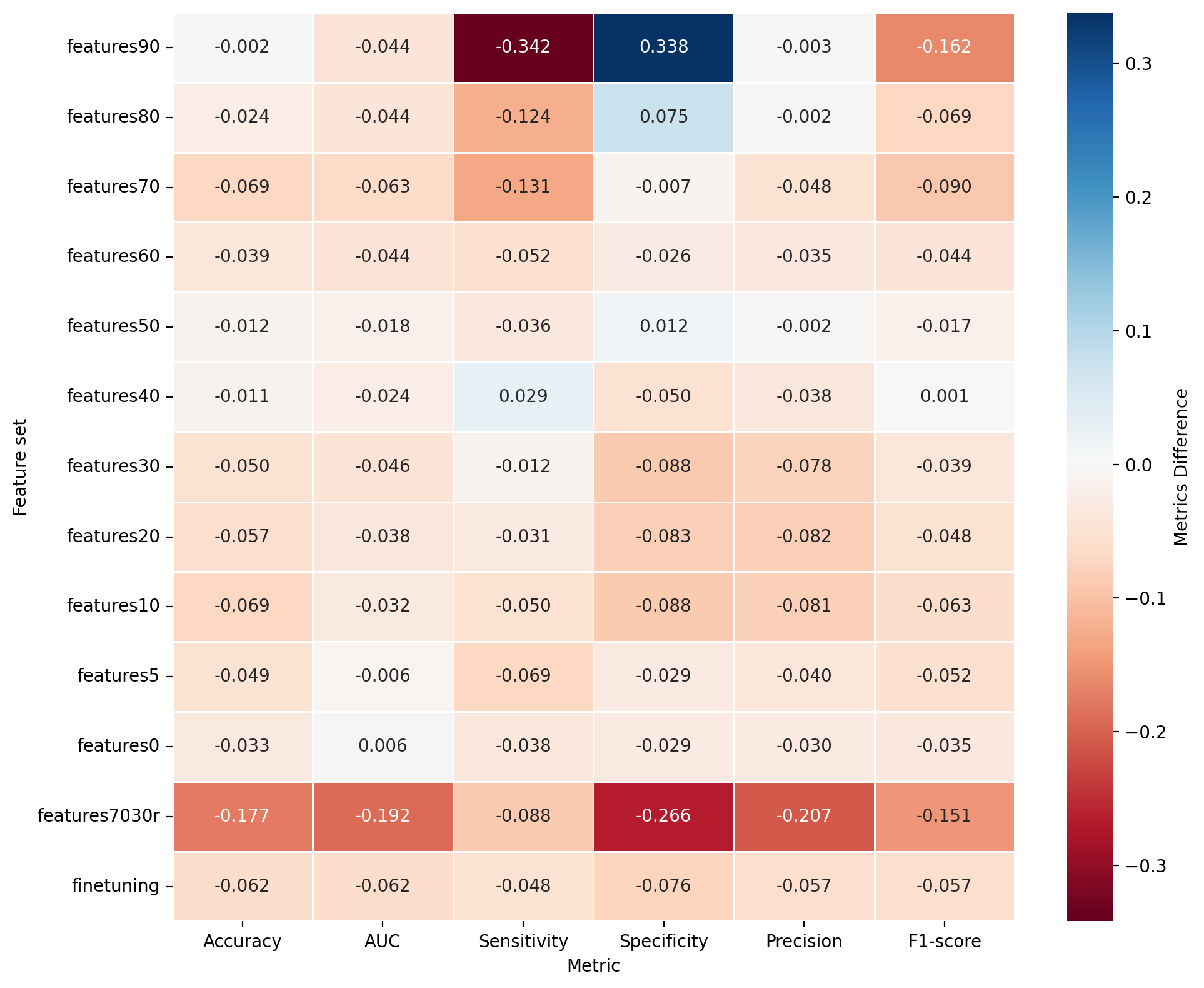}
    \caption{
        Heat‑map of metric differences (PGM $-$ Ensemble) across feature‑selection scenarios. Each row corresponds to a feature‑set,
        and each column represents one of the six evaluation metrics
        \emph{Accuracy}, \emph{AUC}, \emph{Sensitivity},
        \emph{Specificity}, \emph{Precision}, and \emph{F1‑score}.
        Cells are coloured on a diverging blue–red scale centred at zero:
        A blue hue indicates that the PGM classifier outperforms the
        ensemble for that metric (positive difference), whereas a red hue
        denotes inferior performance (negative difference).  The colour
        intensity is proportional to the magnitude of the difference,
        allowing a rapid visual assessment of where the quantum‑inspired
        PGM model provides gains or losses relative to the traditional
        ensemble approach.}
    \label{fig:pgm‑ensemble‑diff‑heatmap}
\end{figure}

% ── Figura Confusion matrices PCa ─────────────────────────────────────────
\begin{figure}[h!]
    \centering
    \includegraphics[width=\textwidth]{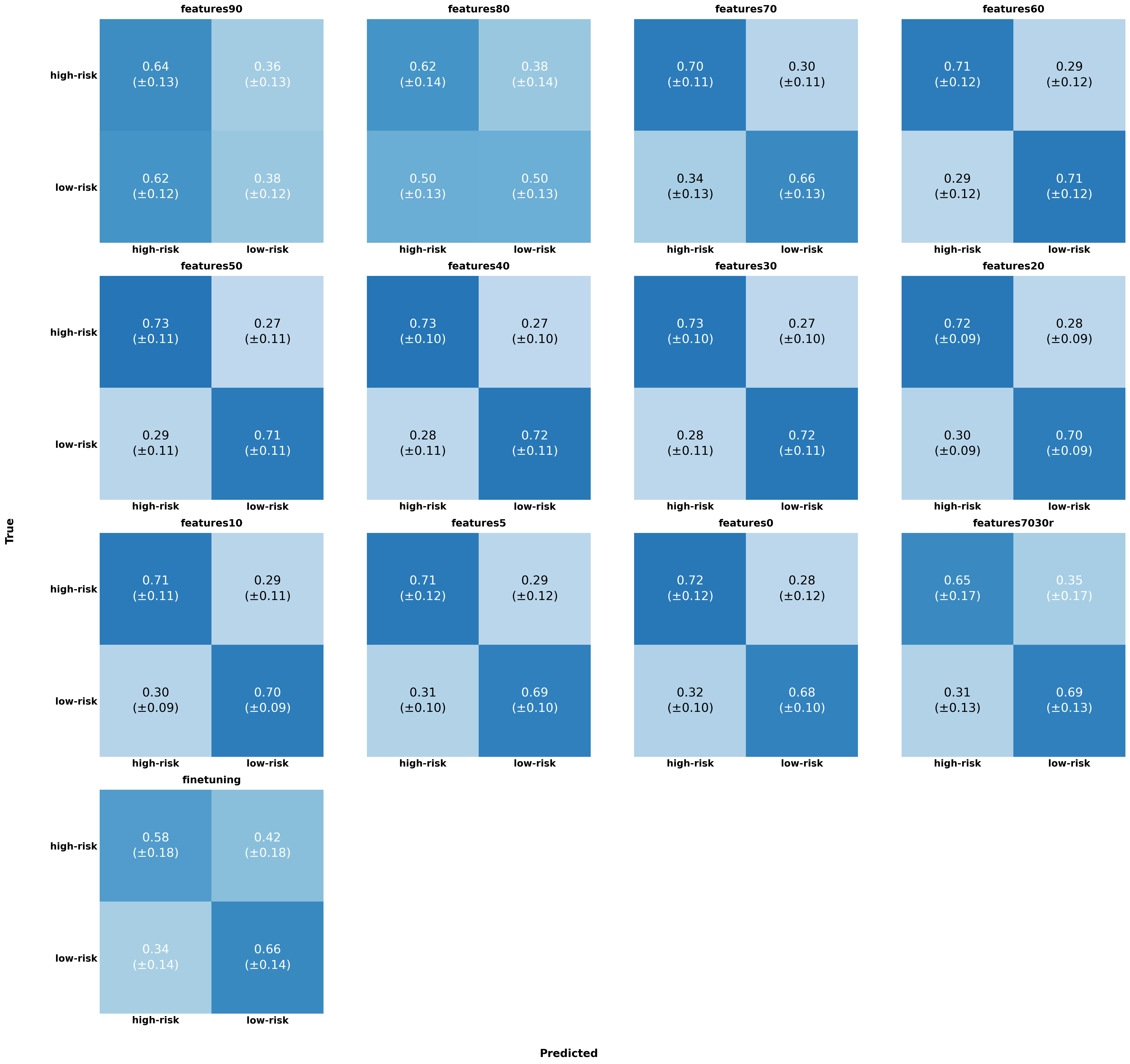}
    \caption{Mean confusion matrices (normalized by row) for the prostate
    cancer (PCa) risk stratification task across the 13 feature subsets,
    averaged over 30 train/test splits. Each cell reports the mean value
    and the standard deviation (±std). The binary classification
    distinguishes high-risk from low-risk patients, using [\textsuperscript{18}F]-PSMA-1007 PET/CT radiomic features extracted from the 13 feature subsets defined by their selection frequency threshold (features90 to features0, finetuning, and
    features7030r).}
    \label{fig:cm_pca}
\end{figure}

\begin{table}[h!]
\centering
\caption{Performance of the PGM model across 13 feature subsets (mean and 95\% CI over 30 splits). Best values per metric are highlighted in \textbf{bold}.}
\label{tab:pgm_pca_subsets}
\resizebox{\textwidth}{!}{%
\begin{tabular}{lcccccc}
\toprule
\textbf{Feature Subset} &
\textbf{Accuracy} &
\textbf{AUC} &
\textbf{Sensitivity} &
\textbf{Specificity} &
\textbf{Precision} &
\textbf{F1-score} \\
\midrule
features90    & 51.35\% (49.26--53.57\%) & 53.55\% (50.61--56.44\%) & 38.42\% (33.74--43.10\%) & 64.29\% (59.85--69.21\%) & 50.71\% (45.37--55.15\%) & 43.07\% (38.64--46.91\%) \\
features80    & 55.67\% (53.45--58.00\%) & 58.09\% (55.43--60.87\%) & 49.75\% (45.07--54.43\%) & 61.58\% (56.90--67.24\%) & 57.51\% (54.00--61.78\%) & 52.21\% (48.84--55.46\%) \\
features70    & 67.98\% (64.53--71.18\%) & 74.96\% (71.02--78.70\%) & 65.52\% (60.84--70.20\%) & 70.44\% (66.26--74.63\%) & 69.01\% (65.52--72.50\%) & 66.78\% (62.85--70.54\%) \\
features60    & 70.94\% (67.49--74.38\%) & 78.34\% (74.91--81.51\%) & 71.18\% (66.74--75.38\%) & 70.69\% (66.01--74.88\%) & 71.24\% (67.77--74.77\%) & 70.78\% (67.03--74.24\%) \\
features50    & 72.02\% (69.16--75.00\%) & 80.10\% (76.95--82.74\%) & 70.95\% (66.90--74.76\%) & 73.10\% (69.29--76.67\%) & 72.94\% (70.04--76.25\%) & 71.54\% (68.76--74.74\%) \\
features40    & \textbf{72.62\%} (70.12--75.12\%) & 80.00\% (77.53--82.42\%) & 72.14\% (68.33--75.95\%) & 73.10\% (69.52--76.43\%) & 73.21\% (70.44--75.92\%) & 72.28\% (69.51--74.97\%) \\
features30    & 72.38\% (69.76--75.24\%) & 79.35\% (76.36--82.18\%) & \textbf{72.14\%} (68.57--75.95\%) & 72.62\% (69.05--75.95\%) & 72.83\% (69.85--75.54\%) & 72.14\% (69.19--74.97\%) \\
features20    & 71.07\% (68.21--73.45\%) & 78.64\% (75.34--81.51\%) & 70.24\% (66.90--73.57\%) & 71.90\% (68.81--75.00\%) & 71.70\% (68.97--74.51\%) & 70.71\% (68.11--73.45\%) \\
features10    & 70.36\% (67.50--73.21\%) & 77.76\% (74.34--80.75\%) & 70.00\% (66.43--73.10\%) & 70.71\% (66.67--74.52\%) & 71.01\% (67.97--74.31\%) & 70.20\% (67.54--73.00\%) \\
features5     & 70.12\% (67.14--73.10\%) & 77.76\% (74.37--80.65\%) & 68.81\% (65.24--72.14\%) & 71.43\% (67.38--75.48\%) & 71.28\% (67.79--74.48\%) & 69.68\% (66.96--72.57\%) \\
features0     & 70.12\% (67.14--72.98\%) & 76.82\% (73.52--79.83\%) & 68.33\% (64.52--71.90\%) & \textbf{71.90\%} (67.62--75.95\%) & 71.57\% (68.18--74.76\%) & 69.47\% (66.61--72.30\%) \\
finetuning    & 61.81\% (57.42--66.07\%) & 66.59\% (61.91--70.88\%) & 65.93\% (59.89--71.15\%) & 57.69\% (50.82--64.56\%) & 62.19\% (57.54--66.80\%) & 63.21\% (59.29--67.14\%) \\
features7030r & 67.02\% (63.21--70.48\%) & \textbf{75.49\%} (71.61--79.13\%) & 68.81\% (63.57--73.33\%) & 65.24\% (58.33--70.71\%) & \textbf{67.39\%} (63.54--71.20\%) & \textbf{67.36\%} (63.19--71.04\%) \\
\bottomrule
\end{tabular}%
}
\end{table}

%\clearpage
\newpage
\subsection{Discussion}

The results reveal two distinct patterns across the considered datasets.

For the NSCLC dataset, the comparison is particularly informative. In the 2-class setting, the PGM classifier consistently improves upon all the classifiers considered in the previous study, showing gains both in terms of accuracy and in terms of ROC-based performance. The same trend is confirmed in the 3-class setting as well, although the advantage is somewhat less pronounced than in the binary case. In contrast, the picture changes in the 4-class setting: while PGM remains competitive, its superiority is no longer observed and it does not emerge as the best-performing classifier among those tested.
We may interpret the decline in performance as the number of classes increases as a natural effect of the greater geometric complexity of the problem. In particular, when moving from 2 to 3–4 classes, it becomes more likely that some classes lie closer to one another — and thus overlap more — in the state space induced by the encoding; consequently, the distributions associated with different classes tend to be less clearly separated, and the discrimination among alternative labels becomes intrinsically more ambiguous, with an expected impact on overall performance. 

Moreover, in the 4-class setting, the increased difficulty may also depend on the clinical nature of the labels: the NOS class is intrinsically heterogeneous and, in addition, it may be radiomically less distinct from LCC. This promotes overlap and increases the risk of misclassification between NOS and LCC. In addition, class imbalance (especially the small sample size of NOS) may amplify split-to-split variability and reduce the stability of per-class estimates.

In this sense, the geometric and clinical explanations should not be regarded as completely independent. The increased overlap in the encoded state space is not merely an abstract consequence of moving to a higher number of classes, but may also reflect the intrinsic clinical and radiomic structure of the problem. In particular, the heterogeneity of the NOS category and its limited radiomic separability from LCC can translate into less clearly separated class distributions after the encoding. Therefore, the reduced geometric separability observed in the 4-class setting appears consistent with the underlying biological and diagnostic ambiguity of the labels, rather than being only a purely mathematical effect of the multiclass formulation.

The PCa dataset exhibits a different behavior. Across versions of the dataset with different numbers of selected features, PGM only rarely achieves the top performance. However, even when it is not the best method, its results are often extremely close to those obtained by the ensemble approach, with differences that are frequently very small. This suggests that, in this experimental setting, PGM provides a solid and stable alternative whose performance remains near the strongest baseline, even when it does not take the lead.

Moreover, the per-metric comparison on the PCa dataset highlights an additional point of particular interest: depending on the feature subset, the difference between PGM and the ensemble can take noticeably different values in terms of sensitivity and specificity. In a clinical risk-stratification scenario, this trade-off is relevant: if the primary goal is to minimize false negatives (high-risk cases classified as low-risk), sensitivity should be prioritized; conversely, if the aim is to reduce invasive procedures and overtreatment, higher specificity may be preferable. This, in turn, suggests the possibility of calibrating the method in a cost-sensitive manner (e.g., by acting on priors or on decision rules), thereby aligning the output with clinical priorities.

\section{Conclusions}

In this work we investigated the application of a quantum-inspired classification strategy based on Pretty Good Measurement to radiomic datasets in two clinically relevant scenarios: lung cancer subtyping and prostate cancer risk stratification. The proposed approach was evaluated under the same experimental conditions adopted in previous studies, allowing for a direct and transparent comparison with established machine learning classifiers.

The experimental analysis does not aim to claim a general superiority of PGM over other methods. Rather, our results show that, in specific settings—most notably in the binary and three-class configurations of the NSCLC dataset—the PGM classifier can achieve clearly competitive, and in some cases improved, performance in terms of both accuracy and ROC-based metrics. In more complex configurations, such as the four-class case or in the PCa dataset, PGM does not consistently outperform the strongest baselines; however, its performance remains very close to that of leading ensemble approaches. 
\textcolor{black}{This consistent competitiveness suggests that, within the benchmark settings considered here, PGM should be regarded as a viable and robust alternative to the strongest classical baselines included in the comparison in radiomics classification tasks.}

More broadly, the present study further supports the practical relevance of quantum-inspired methodologies in biomedical imaging. While previous work had already demonstrated the applicability of this framework in binary classification settings, the results reported here extend its use to multi-class problems, showing that the operator-based encoding and measurement strategy can be effectively generalized beyond the simplest scenarios. In this sense, the paper provides additional evidence that quantum-inspired approaches can offer meaningful and practically useful tools for high-dimensional bioimaging applications. 

Possible future developments follow naturally from what we have observed. A first step is to clarify more systematically in which regimes the PGM approach tends to yield a performance advantage. Our results suggest that this advantage is not universal, but depends sensitively on the geometry of the problem in the representation space induced by the encoding: in other words, on how the samples from each class are arranged after the operator-valued mapping, and on how much the corresponding class distributions are separated or overlapping. From this perspective, it is natural to complement standard classification metrics with a more direct characterization of the induced geometric structure, for instance via distance/similarity measures between class representatives, overlap indices, and proxies for multi-class separability, and to study how these indicators change when varying the encoding, normalization/scaling parameters, and pre-processing choices. Extending the analysis to additional datasets and multi-class settings would thus help delineate more precisely the conditions under which the PGM rule acts as a favorable inductive bias, and would support a more controlled interpretation of the performance trends as the complexity of the task increases.

At the same time, with a view toward using the method as a support tool, it is useful to complement aggregate performance with an analysis of the error structure, i.e., which class confusions are more likely and how relevant they are from an application-driven standpoint. In biomedical domains, different types of errors may have different downstream consequences; accordingly, a natural extension is to introduce a form of cost-sensitive optimization, in which some confusions are penalized more than others. Operationally, this amounts to moving from an evaluation where all errors are treated as equivalent to one in which the procedure is calibrated to reduce the most critical confusions first, thereby making the classifier's output more consistent with contextual constraints while preserving the overall methodological framework.

In conclusion, these research directions aim to strengthen both the understanding of the mechanisms that make PGM competitive in the multi-class setting and the robustness and application alignment of the resulting pipeline. Overall, they can help consolidate PGM as a reliable and flexible quantum-inspired component within machine-learning pipelines for biomedical data, while more precisely identifying the use cases in which it represents a particularly effective option.

\subsection*{Data Availability}

The datasets used in the experiments are provided as supplementary files. The code developed during the experiments is publicly available at the following repository: \href{https://github.com/QuantumUnica/Pretty-Good-Measurement-PGM-Quantum-Inspired-Classifier-for-Radiomic.git}{GitHub}

\section*{Funding declaration}

Giuseppe Sergioli and Roberto Giuntini acknowledge financial support from the PRIN-PNRR project   “\emph{Quantum Models for Logic, Computation and Natural Processes} – Qm4Np” (code: P2022A52CR), from the PRIN-2022 project “\emph{The COst of Reasoning: Theory and EXperiments} – CORTEX” (code: 2022ZLLR3T) and from the Fondazione di Sardegna project “\emph{An algebraic approach to
hyperintensionality}” (code: F23C25000360007). 

Roberto Giuntini is partially funded by the TÜV SÜD Foundation, the Federal Ministry of Education and Research (BMBF) and the Free State of Bavaria under the Excellence Strategy of the Federal Government and the Länder, as well as by the Technical University of Munich-Institute for Advanced Study. 

Alessandro Stefano and Giorgio Russo acknowledge financial support from the project “The Breast Integrated Solution (BIS): technological innovation to support integrated diagnosis and personalized treatment of breast cancer", funded by PR FESR 2021/2027, B.U. No. 30 of 27/07/2023 – D.D. No. 320 of 25/07/2023.

% ------------ APPENDIX -------------------- 
\newpage
\appendix

\section{Radiomics feature extraction parameters}
\label{app:extraction_params}

The radiomics features used in the present work were not extracted ex novo: for both case studies, we relied directly on the feature-selected and harmonized data made available by the authors of the reference studies~\cite{Pasini2023, Pasini2025}, whose extraction protocols are therefore inherited from those works. To assist the reader in interpreting the radiomics inputs to the PGM classifier without the need to consult both reference papers, this appendix consolidates the corresponding PyRadiomics configurations in a single side-by-side summary. Parameters not listed are left at their PyRadiomics default values.

\begin{table}[h!]
\centering
\caption{PyRadiomics feature extraction configurations of the two reference studies whose feature-selected and harmonized data were used in the present work. The NSCLC (CT) and PCa ($[^{18}$F$]$PSMA-1007 PET) configurations differ in most parameters, reflecting the distinct imaging modalities and the independent protocols established in the respective reference studies. Parameters not listed are left at their PyRadiomics default values. NSCLC configuration adapted from Table~A1 of~\cite{Pasini2023}; PCa configuration adapted from Table~1 of~\cite{Pasini2025}.}
\label{tab:appendix_extraction}
\begin{tabular}{lll}
\toprule
\textbf{Parameter} & \textbf{NSCLC (CT)} & \textbf{PCa ($[^{18}$F$]$PSMA-1007 PET)} \\
\midrule
Bin discretization        & Bin count = 64                & Bin width = 0.25 \\
Isotropic voxel size      & $1\times1\times1$ mm$^3$      & $2\times2\times2$ mm$^3$ \\
Interpolator              & sitkLinear                    & sitkBSpline \\
Wavelet basis             & Haar                          & Coiflet 1 \\
LoG sigma range           & 0.5--5.0, step 0.5            & 0.5--5.0, step 0.5 \\
Image normalization       & --                            & Enabled (scale = 1) \\
Intensity standardization & Hounsfield Units (CT)         & SUV$_{\mathrm{bw}}$ (g/ml) \\
\bottomrule
\end{tabular}
\end{table}

\newpage

\section{Detailed specification of the feature subsets used in the PCa experiments}
\label{app:feature_subsets}

As detailed in Section~\ref{sec:ex_protoc}, the 13 feature subsets used in the prostate cancer (PCa) experiments were not constructed by us: they were derived from the per-split, post-LASSO feature lists made available by the authors of~\cite{Pasini2025} for each of the 30 stratified splits, by applying the selection-frequency thresholds and the correlation criterion originally proposed in that work. To assist the reader in interpreting our PCa results without requiring a parallel consultation of~\cite{Pasini2025}, this appendix collects in a single, self-contained reference the construction criterion and the size of each subset (Table~\ref{tab:appendix_subsets}), and the explicit composition of the \textit{finetuning} subset together with the corresponding average Pearson Correlation Coefficient (PCC) values with respect to \textit{features70} (Table~\ref{tab:appendix_pcc}). Features with an average PCC below the $0.3$ threshold (highlighted in bold) are those retained in the final \textit{finetuning} subset. For maximum reproducibility of the eleven frequency-based subsets, Table~\ref{tab:appendix_all_features} additionally reports the selection frequency of all 79 features that were retained at least once across the 30 LASSO iterations, from which the composition of any frequency-based subset (\textit{features90}, \textit{features80}, \ldots, \textit{features0}) can be directly reconstructed by applying the corresponding threshold of Table~\ref{tab:appendix_subsets}.

\begin{table}[h!]
\centering
\footnotesize
\caption{Specification of the 13 feature subsets used to train the PGM classifier on the PCa dataset. The feature-frequency thresholds and subset sizes follow the construction proposed in~\cite{Pasini2025} (adapted from Table~4 therein). The \textit{features7030r} subset is defined as the union of \textit{finetuning} and \textit{features70}.}
\label{tab:appendix_subsets}
\begin{tabular}{lp{9cm}c}
\toprule
\textbf{Subset name} & \textbf{Selection criterion} & \textbf{Size} \\
\midrule
features90     & feature frequency $\geq 90\%$ & 1 \\
features80     & feature frequency $\geq 80\%$ & 2 \\
features70     & feature frequency $\geq 70\%$ & 4 \\
features60     & feature frequency $\geq 60\%$ & 5 \\
features50     & feature frequency $\geq 50\%$ & 8 \\
features40     & feature frequency $\geq 40\%$ & 12 \\
features30     & feature frequency $\geq 30\%$ & 16 \\
features20     & feature frequency $\geq 20\%$ & 23 \\
features10     & feature frequency $\geq 10\%$ & 34 \\
features5      & feature frequency $\geq 5\%$  & 52 \\
features0      & all selected features         & 79 \\
finetuning     & features with $30\% \leq$ frequency $< 70\%$ and average PCC $< 0.3$ w.r.t. \textit{features70} & 7 \\
features7030r  & union of \textit{finetuning} and \textit{features70} & 11 \\
\bottomrule
\end{tabular}
\end{table}

\begin{table}[h!]
\centering
\footnotesize
\caption{Candidate features for the \textit{finetuning} subset, i.e., features with a selection frequency in the $[30\%, 70\%)$ range. For each feature, the average Pearson Correlation Coefficient (PCC) with respect to the features in the \textit{features70} subset (averaged over the 30 iterations of the preliminary pipeline) and the selection frequency are reported. Features in bold are those satisfying $\mathrm{PCC} < 0.3$ and therefore retained in the final \textit{finetuning} subset. Adapted from Table~5 of~\cite{Pasini2025}.}
\label{tab:appendix_pcc}
%\begin{tabular}{lcc}
\begin{tabular}{p{8cm}rr}
\toprule
\textbf{Feature} & \textbf{Average PCC} & \textbf{Feature frequency} \\
\midrule
log-sigma-4--5-mm-3D-glcm-InverseVariance                                & 0.3369 & 60.00\% \\
\textbf{log-sigma-0--5-mm-3D-ngtdm-Busyness}                             & \textbf{0.2735} & \textbf{56.67\%} \\
log-sigma-3--0-mm-3D-glszm-SmallAreaEmphasis                             & 0.3756 & 53.33\% \\
wavelet-LHL-glcm-MaximumProbability                                      & 0.3758 & 50.00\% \\
log-sigma-5--0-mm-3D-glszm-GrayLevelNonUniformityNormalized              & 0.3322 & 46.67\% \\
\textbf{log-sigma-2--5-mm-3D-glszm-GrayLevelNonUniformity}               & \textbf{0.0948} & \textbf{43.33\%} \\
\textbf{wavelet-HHL-gldm-SmallDependenceLowGrayLevelEmphasis}            & \textbf{0.1595} & \textbf{43.33\%} \\
\textbf{wavelet-HLH-glszm-GrayLevelNonUniformity}                        & \textbf{0.1341} & \textbf{43.33\%} \\
\textbf{wavelet-HLH-glcm-Idn}                                            & \textbf{0.1833} & \textbf{36.67\%} \\
\textbf{wavelet-HHL-glcm-Idn}                                            & \textbf{0.2470} & \textbf{33.33\%} \\
wavelet-HLL-gldm-DependenceVariance                                      & 0.3301 & 33.33\% \\
\textbf{log-sigma-4--0-mm-3D-firstorder-Skewness}                        & \textbf{0.1659} & \textbf{30.00\%} \\
\bottomrule
\end{tabular}
\end{table}

\begin{table}[H]
\centering
\scriptsize
\caption{Selection frequency of all 79 features that were selected at least once across the 30 LASSO iterations of the preliminary pipeline of~\cite{Pasini2025}. For each feature, the selection frequency is reported as the percentage of iterations in which the feature was retained by LASSO. Features are sorted by decreasing frequency. The eleven frequency-based subsets defined in Table~\ref{tab:appendix_subsets} (\textit{features90}, \textit{features80}, \textit{features70}, \ldots, \textit{features5}, \textit{features0}) correspond to the cumulative groups of features whose frequency satisfies the respective threshold; for instance, \textit{features70} consists of the four features with frequency $\geq 70\%$ (top of the table), and \textit{features0} comprises all 79 listed features. Computed by the present authors from the per-split, post-LASSO feature lists made available by the authors of~\cite{Pasini2025}; consistent with Figure~4 and Table~4 of~\cite{Pasini2025}.}
\label{tab:appendix_all_features}
\begin{tabular}{p{10cm}c}
\toprule
\textbf{Feature} & \textbf{Selection frequency} \\
\midrule
wavelet-LHH-firstorder-Skewness & 96.67\% \\
wavelet-HHL-glszm-SmallAreaEmphasis & 86.67\% \\
wavelet-LHL-gldm-DependenceVariance & 76.67\% \\
wavelet-HHL-glszm-SmallAreaLowGrayLevelEmphasis & 73.33\% \\
log-sigma-4-5-mm-3D-glcm-InverseVariance & 60.00\% \\
log-sigma-0-5-mm-3D-ngtdm-Busyness & 56.67\% \\
log-sigma-3-0-mm-3D-glszm-SmallAreaEmphasis & 53.33\% \\
wavelet-LHL-glcm-MaximumProbability & 50.00\% \\
log-sigma-5-0-mm-3D-glszm-GrayLevelNonUniformityNormalized & 46.67\% \\
log-sigma-2-5-mm-3D-glszm-GrayLevelNonUniformity & 43.33\% \\
wavelet-HLH-glszm-GrayLevelNonUniformity & 43.33\% \\
wavelet-HHL-gldm-SmallDependenceLowGrayLevelEmphasis & 43.33\% \\
wavelet-HLH-glcm-Idn & 36.67\% \\
wavelet-HLL-gldm-DependenceVariance & 33.33\% \\
wavelet-HHL-glcm-Idn & 33.33\% \\
log-sigma-4-0-mm-3D-firstorder-Skewness & 30.00\% \\
wavelet-LLL-glszm-LargeAreaLowGrayLevelEmphasis & 26.67\% \\
wavelet-HLH-glcm-Idmn & 26.67\% \\
log-sigma-3-5-mm-3D-glszm-GrayLevelNonUniformity & 23.33\% \\
wavelet-HLL-glszm-ZoneEntropy & 23.33\% \\
wavelet-HLH-glcm-Correlation & 20.00\% \\
original-glrlm-ShortRunLowGrayLevelEmphasis & 20.00\% \\
log-sigma-3-5-mm-3D-glcm-InverseVariance & 20.00\% \\
original-firstorder-10Percentile & 13.33\% \\
wavelet-LHH-glcm-ClusterShade & 13.33\% \\
log-sigma-3-0-mm-3D-glszm-SmallAreaLowGrayLevelEmphasis & 13.33\% \\
wavelet-HLL-glszm-GrayLevelNonUniformity & 13.33\% \\
wavelet-HLH-glszm-SmallAreaEmphasis & 10.00\% \\
wavelet-HHH-glszm-LowGrayLevelZoneEmphasis & 10.00\% \\
log-sigma-4-5-mm-3D-firstorder-Skewness & 10.00\% \\
wavelet-HLL-glcm-Idn & 10.00\% \\
log-sigma-1-5-mm-3D-gldm-DependenceVariance & 10.00\% \\
wavelet-HLL-firstorder-Uniformity & 10.00\% \\
log-sigma-4-0-mm-3D-glszm-LargeAreaLowGrayLevelEmphasis & 10.00\% \\
log-sigma-0-5-mm-3D-glszm-GrayLevelNonUniformity & 6.67\% \\
log-sigma-4-5-mm-3D-glszm-SmallAreaEmphasis & 6.67\% \\
wavelet-LHH-glcm-InverseVariance & 6.67\% \\
wavelet-HLL-gldm-LargeDependenceEmphasis & 6.67\% \\
wavelet-HLL-glcm-MaximumProbability & 6.67\% \\
log-sigma-2-0-mm-3D-glszm-GrayLevelNonUniformity & 6.67\% \\
wavelet-LLH-ngtdm-Coarseness & 6.67\% \\
wavelet-HHH-gldm-DependenceVariance & 6.67\% \\
log-sigma-4-5-mm-3D-glszm-GrayLevelNonUniformityNormalized & 6.67\% \\
wavelet-LLL-glszm-LargeAreaHighGrayLevelEmphasis & 6.67\% \\
wavelet-LLL-glszm-SizeZoneNonUniformityNormalized & 6.67\% \\
log-sigma-1-0-mm-3D-glszm-SmallAreaEmphasis & 6.67\% \\
log-sigma-3-0-mm-3D-glszm-GrayLevelNonUniformity & 6.67\% \\
wavelet-LHL-glcm-Correlation & 6.67\% \\
log-sigma-0-5-mm-3D-glszm-LowGrayLevelZoneEmphasis & 6.67\% \\
log-sigma-0-5-mm-3D-gldm-LargeDependenceLowGrayLevelEmphasis & 6.67\% \\
wavelet-HLL-firstorder-InterquartileRange & 6.67\% \\
log-sigma-0-5-mm-3D-glszm-LargeAreaLowGrayLevelEmphasis & 6.67\% \\
log-sigma-0-5-mm-3D-glrlm-LongRunLowGrayLevelEmphasis & 3.33\% \\
log-sigma-2-5-mm-3D-glcm-MaximumProbability & 3.33\% \\
wavelet-HLH-ngtdm-Busyness & 3.33\% \\
original-glszm-SmallAreaLowGrayLevelEmphasis & 3.33\% \\
wavelet-HLL-firstorder-Skewness & 3.33\% \\
wavelet-HLL-glcm-JointEnergy & 3.33\% \\
wavelet-HLL-glcm-Idmn & 3.33\% \\
log-sigma-3-5-mm-3D-firstorder-Skewness & 3.33\% \\
wavelet-LHL-firstorder-Skewness & 3.33\% \\
wavelet-HLH-glszm-LargeAreaHighGrayLevelEmphasis & 3.33\% \\
log-sigma-3-0-mm-3D-glszm-LargeAreaHighGrayLevelEmphasis & 3.33\% \\
wavelet-HHL-glcm-Idmn & 3.33\% \\
log-sigma-1-0-mm-3D-glszm-SmallAreaLowGrayLevelEmphasis & 3.33\% \\
log-sigma-2-5-mm-3D-glszm-SmallAreaEmphasis & 3.33\% \\
wavelet-LHL-firstorder-Kurtosis & 3.33\% \\
wavelet-LLL-firstorder-Median & 3.33\% \\
wavelet-HHH-glcm-MaximumProbability & 3.33\% \\
log-sigma-0-5-mm-3D-gldm-LowGrayLevelEmphasis & 3.33\% \\
log-sigma-0-5-mm-3D-glszm-SmallAreaEmphasis & 3.33\% \\
wavelet-HHL-firstorder-Median & 3.33\% \\
wavelet-HHH-glszm-GrayLevelNonUniformityNormalized & 3.33\% \\
log-sigma-5-0-mm-3D-gldm-LargeDependenceHighGrayLevelEmphasis & 3.33\% \\
wavelet-HLL-firstorder-Energy & 3.33\% \\
log-sigma-3-5-mm-3D-glszm-ZoneEntropy & 3.33\% \\
log-sigma-5-0-mm-3D-firstorder-Skewness & 3.33\% \\
wavelet-HHL-glcm-InverseVariance & 3.33\% \\
log-sigma-4-0-mm-3D-glszm-SmallAreaLowGrayLevelEmphasis & 3.33\% \\
\bottomrule
\end{tabular}
\end{table}

\printbibliography
\end{document}